\documentclass[preprint,12pt,authoryear]{elsarticle}
\usepackage{graphicx}
\usepackage{color}
\usepackage{caption}
\usepackage{subcaption}
\usepackage{subfloat}
\usepackage{multirow}
\usepackage{amsmath}
\usepackage{bm}
\usepackage{enumerate}
\usepackage{pdflscape}
\usepackage{booktabs,caption,fixltx2e}
\usepackage[flushleft]{threeparttable}
\usepackage{placeins}
\date{}
\usepackage{amsfonts}
\usepackage{url}
\usepackage{setspace}
\usepackage{hyperref}
\usepackage{bbm}
\usepackage[margin=1in]{geometry}
\usepackage{booktabs}
\usepackage{algpseudocode}
\usepackage{pbox}
\usepackage[lined,boxed,commentsnumbered]{algorithm2e}
\usepackage{verbatimbox}

\allowdisplaybreaks
\usepackage{enumitem}
\usepackage{natbib}
\usepackage[labelformat=simple]{subcaption}

\usepackage{bbm}
\usepackage{natbib}
\usepackage{epstopdf}
\usepackage{MnSymbol,wasysym}
\usepackage{mwe,tikz}\usepackage[percent]{overpic}
\usepackage{tabularx}
\usepackage{booktabs}
\usepackage[export]{adjustbox}

\usepackage{xcolor}

\usepackage{afterpage}

\newtheorem{remark}{Remark}

\usepackage{tikz}
\usetikzlibrary{positioning,shapes}
\usepackage{soul} 
\usepackage{xcolor}
\usepackage{afterpage}
\usepackage{capt-of}

\journal{Transportation Research Part C} 

\begin{document}

	\begin{frontmatter}
	\title{Estimating Historical Hourly Traffic Volumes via Machine Learning and Vehicle Probe Data: A Maryland Case Study\footnote[2]{An earlier version of this paper \citep{sekula2017applications} was presented at a TRB annual meeting.}}
	
	\author[label1,label2]{Przemys\l{}aw Seku\l{}a}
	
	\author[label1]{Nikola Markovi\'c}		
	
	\author[label1]{Zachary Vander Laan}
	
	\author[label1]{Kaveh Farokhi Sadabadi}
	
	\address[label1]{%Center for Advanced Transportation Technology, 
		Department of Civil and Environmental Engineering, University of Maryland, College Park, MD, USA}
	
	\address[label2]{Faculty of Informatics and Communication, University of Economics in Katowice, Katowice, Poland}

	\begin{abstract}
		This paper focuses on the problem of estimating historical traffic volumes between sparsely-located traffic sensors, which transportation agencies need to accurately compute statewide performance measures. To this end, the paper examines applications of vehicle probe data, automatic traffic recorder counts, and neural network models to estimate hourly volumes in the Maryland highway network, and proposes a novel approach that combines neural networks with an existing profiling method. On average, the proposed approach yields 24\% more accurate estimates than volume profiles, which are currently used by transportation agencies across the US to compute statewide performance measures. The paper also quantifies the value of using vehicle probe data in estimating hourly traffic volumes, which provides important managerial insights to transportation agencies interested in acquiring this type of data. For example, results show that volumes can be estimated with a mean absolute percent error of about 21\% at locations where average number of observed probes is between 30 and 47 vehicles/hr, which provides a useful guideline for assessing the value of probe vehicle data from different vendors.
	\end{abstract}
	
	\begin{keyword}
		traffic volume estimation \sep regression \sep multi-layered neural networks \sep vehicle probe data
	\end{keyword}
	
\end{frontmatter}

\section{Introduction}

Hourly traffic volumes and speeds are important inputs needed by transportation agencies for calculating statewide performance measures. While average vehicle speeds can be readily obtained across the entire state road network (e.g., via probe vehicles), traffic volumes are particularly challenging to determine, as vehicle counts are typically only available at a handful of locations where fixed traffic sensors are located. Currently, many transportation agencies apply speed profiles or other factors to average annual daily traffic (AADT) volumes to obtain reasonable hourly volume estimates \citep{schrank20152015}. However, these hourly profiles do not reflect variations in traffic due to weather, incidents, or daily demand fluctuations, and it is often impractical and expensive to install more sensors \citep{ITSpaper2017}. Thus, in order to perform more detailed analyses that require accurate volume data, transportation agencies must find ways to improve hourly volume estimates based on limited data sources.

Existing approaches for estimating traffic volumes span multiple research areas, and are briefly discussed in order to articulate the problem addressed in this paper. Broadly speaking, volume estimation research can be divided into (a) estimating \underline{historical} and (b) predicting \underline{future} volumes.
\begin{itemize}
	\item \textit{Estimating historical volumes}. Much of the existing work in this area is concerned with estimating AADT data, which represents a very basic measure of travel demand \citep{aashto2001policy}. A number of approaches for estimating AADT via regression have been proposed in the literature, encompassing both statistical \citep{mohamad1998annual,zhao2004using,kingan2006robust} and machine learning techniques \citep{sharma2001application, castro2009aadt, islam2016estimation}. For example, \cite{islam2016estimation} applies artificial neural networks (ANN) and support vector machines (SVM) to estimate AADT based on road geometry, existing counts and local socio-economic data. Related applications of SVM and regression are discussed in \cite{castro2009aadt} and \cite{zhao2001contributing}, respectively. Furthermore, a technique to transform AADT into hourly volume profiles for a typical week is proposed in \cite{schrank20152015}, and is widely used in practice to obtain hourly volume estimates that are needed to compute historical network-wide performance measures (e.g., see \citealt{schrank2012tti,subrat2015}). Another historical volume estimation approach involves utilizing macroscopic traffic models \citep{papageorgiou2010traffic}, often in conjunction with data assimilation techniques such as the Kalman filtering \citep{evensen2009data}, examples of which include \cite{herrera2007traffic}, \cite{work2008ensemble} and \cite{blandin2012sequential}. While this approach has shown promise on small networks, it is not scalable to the statewide level.
	
	\item \textit{Predicting future volumes}. This line of research encompasses short-term and long-term volume predictions, with the vast majority dealing with the former approach.  Short-term prediction generally involves building models or learning patterns from a series of historical volume measurements, and uses them to infer volumes for a time interval in the immediate future. A comprehensive review of various short-term traffic prediction methods is provided in \cite{vlahogianni2014short}, which categorizes each approach as time series analysis, function approximation, optimization, pattern recognition or clustering.  These approaches can also be categorized as either parametric or non-parametric, with common parametric approaches including variants of the autoregressive integrated moving average (ARIMA) model (e.g., \citealt{levin1980forecasting}; \citealt{lee1999application}; \citealt{williams2003modeling}), other non-ARIMA time series models (e.g., \citealt{yu2003short,ghosh2007bayesian}), state-space models (e.g., \citealt{stathopoulos2003multivariate}), and non-parametric techniques including $k$-nearest neighbor (e.g.,  \citealt{gong2002three}), support vector regression (e.g., \citealt{zhang2009traffic}), and neural networks (e.g., \citealt{vlahogianni2005optimized}). A small body of recent non-parametric research in this area includes applications of deep learning, examples of which are \cite{lv2015traffic} and \cite{polson2017deep}.  On the other hand, long-term traffic prediction uses historical measurements to predict volumes farther into the future, generally at a much less granular level. An example of this approach is \cite{kingan2006robust}, which applies robust regression to forecast traffic volumes several years into the future.  
\end{itemize}

The current paper is concerned with the problem of estimating \underline{historical hourly} traffic volumes, which transportation agencies need for annual computations of various network-wide performance measures (e.g., user delay cost, energy efficiency). The paper makes two primary contributions:
\begin{itemize}	
	\item It proposes a novel approach for estimating hourly volumes that combines a widely-used profiling method \citep{schrank20152015} and an ANN model trained with vehicle probe data from one of the leading GPS companies in North America. The proposed approach yields 24\% more accurate volume estimates than the sole profiling method, which represents the state-of-the-practice. Accordingly, the proposed approach could substantially improve the operations of numerous transportation agencies that are currently relying on hourly profiles to compute various performance measures. The prerequisite for implementing the proposed approach would be acquisition of vehicle probe data, whose value in estimating volumes is examined as the second major contribution of this work. 
	
	\item It quantifies the value of using vehicle probe data in estimating hourly traffic volumes, which provides important managerial insights to transportation agencies interested in acquiring this type of data. For example, results show that volumes can be estimated with mean absolute percent error of about 21\% at locations where average number of observed probes is between 30 and 47 vehicles/hr, which provides a useful guideline in assessing the value of probe vehicle data from different vendors. These conclusions are drawn from analysis of a rich probe vehicle dataset that captures between 0.8\% and 4.5\% of Maryland traffic. 	
	
\end{itemize}

The rest of the paper is organized into four sections. First, the data used for the analysis are presented. It is followed by a description of the proposed approach and the procedure used to calibrate and evaluate it. After discussing results and assessing performance of the model, conclusions are drawn and several extensions of this work are proposed.

\section{Data}\label{DataSection}
The basic idea behind the proposed approach is to train an ANN model to learn the relation between traffic volumes and various influencing factors, which would then enable transportation agencies to apply the model and estimate volumes at locations where traffic sensors are unavailable. Accordingly, the proposed work regresses hourly volumes from automatic traffic recording (ATR) stations to explanatory variables obtained from various data sources (e.g., probe vehicles, weather stations). The remainder of this section provides a brief description of the data used to train the model.
\begin{itemize}
	\item \textit{ATR data}. The state of Maryland has a number of ATR stations, which provide aggregate hourly vehicle counts. Data from 45 stations (90 one-directional carriageways) are used as a ground-truth to train and evaluate the performance of the ANN model. The 45 ATR stations are located along different types of roads, including Interstates, US routes, and MD routes.
	
	\item \textit{Vehicle probe volumes}. Raw GPS probe data from one of the leading GPS companies provide a sample of traffic throughout the state of Maryland. Through extensive processing of the raw GPS data described in \cite{markovic2017applications}, 30 min vehicle probe volumes were computed for three different types of vehicles (less than 14k lb, between 14k and 26k lb, above 26k lb) for all the road links in Maryland, including the 45 links with ATR stations that will be used to train and evaluate the ANN model. In total, 9 features were extracted from this dataset and used to estimate volumes for the observed hour. Specifically, for each of the three vehicle weight classes, we consider: (a) vehicle probe volumes during the first 30 minutes of the observed hour, (b) vehicle probe volumes during the second 30 minutes of the observed hour, and (c) vehicle probe volumes during the 30 minutes before the observed hour. It is noteworthy that a comparison of vehicle probe and ATR volumes indicates that the average penetration rate of probe vehicles is about 1.8\%.
	
	\item \textit{Vehicle probe speeds}. The vehicle probe speeds estimated based on GPS data were obtained from the Regional Integrated Transportation Information System \citep{RITISref}, which is a data sharing repository created and maintained by the CATT Lab at the University of Maryland. These data are available throughout the state of Maryland, including the 45 roads with ATR stations that will be used in numerical experiments. Specifically, two speed features were used to estimate hourly traffic volumes: the average hourly speed and the approximate free-flow speed.
	
	\item \textit{Weather data}. Hourly weather data was obtained from \cite{weatherWeb}, and used to extract the following records: Temperature, Visibility, Precipitation, Weather Description (Clear, Mostly Cloudy, Overcast, Scattered Clouds, Partly Cloudy, Unknown, Thunderstorm, Light Rain, Light Snow, Light Drizzle, Rain, Heavy Rain, Squalls, Haze, Freezing Rain, Light Freezing Rain, Drizzle, Light Thunderstorms and Rain, Heavy Thunderstorms and Rain, Thunderstorms and Rain, Mist, Fog, Light Freezing Drizzle, Light Freezing Fog, Heavy Drizzle, Light Thunderstorms and Snow, Snow, Blowing Snow, Heavy Snow, Shallow Fog, Ice Pellets, Patches of Fog, Light Ice Pellets). One-hot encoding of the categorical Weather Description resulted in 36 weather features, which are available throughout the state, including the 45 locations that will be used to evaluate the ANN model.
	
	\item \textit{Infrastructure data}. Road characteristics were obtained from \cite{googleMap} and OpenStreetMap. Specifically, we considered the number of lanes, speed limits, class of the road (motorway or trunk), and type of the road (Interstate, US road or MD road). With one-hot encoding of categorical variables, a total of 7 features were considered for each road link. Although time-consuming to obtain, infrastructural characteristics are available throughout the observed road network, including the 45 locations that will be used in the numerical experiments. 
	
	\item \textit{Temporal data}. Information about the hour of the day (1,..,24) and whether the observed day is Saturday, Sunday or a Federal holiday (Washington's Birthday, Independence Day and Columbus Day that take place during the observed 4 month period), were also considered for each data point in order to account for temporal traffic patterns. With one-hot encoding, these temporal characteristics translated to 29 features.
	
	\item \textit{Volume profiles}. Hourly volume profiles for a typical week were derived by applying the commonly-used profiling method \citep{schrank20152015}, which transforms AADT estimates from the Highway Performance Monitoring System \citep{HPMSweb} into hourly volume profiles based on historical vehicle probe speeds available in RITIS. These estimates were used as another feature in the considered problem of estimating historical hourly volumes.
\end{itemize}

In total, the dataset used in the current analysis includes \ul{200,490 data points with 84 features}. The number of points corresponds to the number of hourly volume measurements at 45 ATR stations, each of which counts traffic in both directions. On the other hand, the features correspond to the explanatory variables listed above (i.e., everything except the ATR data, which is used as the target for training the ANN model). 

\section{Method}
This section provides a brief overview of the proposed approach, which would enable other researchers to replicate the analysis. The following subsection describes a fully-connected ANN with an Exponential Linear Unit (ELU) activation function and a dropout method, which is used to estimate historical hourly volumes. The subsequent subsection explains the leave-one-ATR-out cross-validation and model evaluation approach.  

\subsection{Fully-connected feedforward multi-layer ANN}
A fully-connected feedforward multi-layer ANN consists of neurons that are organized into layers, with the first representing the input variables, the last representing model predictions, and ones in between called hidden layers. The ``fully-connected" and ``feedforward" terms correspond to the structure of connections among neurons; in a fully-connected feedforward multi-layer ANN, each neuron is linked with all the neurons from the previous layer (i.e., no loops). Based on this architecture, model estimates are generated using forward propagation, where the output from a neuron is computed as
\begin{align}
a_i^{(l+1)} = f\left(w_i^{(l+1)}a^{(l)} + b_i^{(l+1)}\right),
\end{align}
and $a_i^{(l+1)}$ denotes the output from the $i$-th neuron in layer $l+1$, $a^{(l)}$ represents the output vector from the neurons in layer $l$, $w_i^{(l+1)}$ is a vector of weights between the $i$-th neuron in layer $l+1$ and all the neurons in layer $l$, $b_i^{(l+1)}$ represents the bias unit associated with the $i$-th neuron in layer $l+1$, and $f$ denotes an activation function used to capture nonlinear relationships.

The ``multi-layer'' term refers to the fact that the model has multiple hidden layers, whereas standard ``shallow" networks have only one. These additional layers can be advantageous; networks with many hidden layers (i.e., deep networks) can capture more complex relationships, and in many situations are able to do so with fewer parameters than adding the neurons to the same hidden layer. However, there are two practical problems with deep networks: overfitting and computational efficiency. Overfitting occurs when the model has too many parameters and learns the patterns in a training dataset without being able to reliably generalize the results to other datasets; it achieves much higher estimation accuracy on training data as compared to cross-validation or testing data.  Common approaches for dealing with this problem are $L_1$ and $L_2$ regularization, but these methods are often insufficient for deep networks with many parameters, thus limiting the number of parameters that can be used and hindering the model's ability to capture subtle dependencies. To keep the network complex enough to capture subtle relations while simultaneously preventing the ANN from overfitting the data, a recently-proposed dropout procedure \citep{Hinton2012dropout} is employed. In dropout, during each training step, randomly selected neurons are ignored, which means that their contribution to the activation of downstream neurons is temporally removed on the forward pass and any weight updates are not applied to the neuron in the backward pass. With dropout, the feedforward operations are defined as
\begin{align}
r^{(l)} &\sim \operatorname{Bern} \left({p}\right),\\
\tilde{a}^{(l)} &= r^{(l)} a^{(l)},\\
a_i^{(l+1)} &= f(w_i^{(l+1)}\tilde{a}^{(l)} + b_i^{(l+1)}),
\end{align}
where $p$ denotes the parameter of the Bernoulli distribution. In this paper, the efficiency of dropout was confirmed during the cross-validation training and testing process. Namely, the error metrics for training and testing datasets were similar, indicating that the overfitting was appropriately addressed.

The second problem with deep networks is computational efficiency of the training procedure. In shallow networks, the most popular choice for an activation function is the sigmoid function,
\begin{align}
f(x)=\left(1+e^{-\lambda x}\right)^{-1},
\end{align}
which has many advantages (e.g., its range is between 0 and 1, and its derivative is easy to compute). However, the derivative of the sigmoid function is between -0.25 and 0.25, and therefore the backpropagation process slows down with each additional layer. To overcome this issue, a recently-introduced ELU activation function \cite{clevert2015fast},
\begin{align}
F(x) =
\begin{cases}
x, & \text{$x > 0$} \\
\alpha \left(e^x-1\right), & \text{$x \le 0$}, \\
\end{cases}       
\end{align}
is employed in this work. For $x>0$ the derivative of ELU is equal to 1, while for $x<0$, $F'(x) = F(x)+\alpha$. It results in faster learning compared to standard activation functions (e.g., sigmoid and tanh), as well as functions for deep learning (e.g., rectified linear unit and leaky rectified linear unit). 

Finally, the ANN is trained using the Adaptive Moment Estimation (Adam) algorithm \citep{kingma2014adam}, which is a recently-introduced stochastic gradient-based optimization method that handles high-dimensional parameter spaces in non-convex optimization problems well. Recent research also shows that Adam outperforms other stochastic gradient-based optimization methods \citep{Ruder2016Optimization} due to its built-in bias-correction.

\begin{remark}
The previously-described ANN encompasses recent advances in training algorithms applicable to large data sets (i.e., Adam \citep{kingma2014adam}), procedures to prevent overfitting (i.e., dropout \citep{Hinton2012dropout}), and best practices in combining the two which are discussed in \citep{srivastava2014dropout}.
\end{remark}

\subsubsection{Hyperparameters}
\begin{figure}
	\centering
	\includegraphics[width=0.9\textwidth]{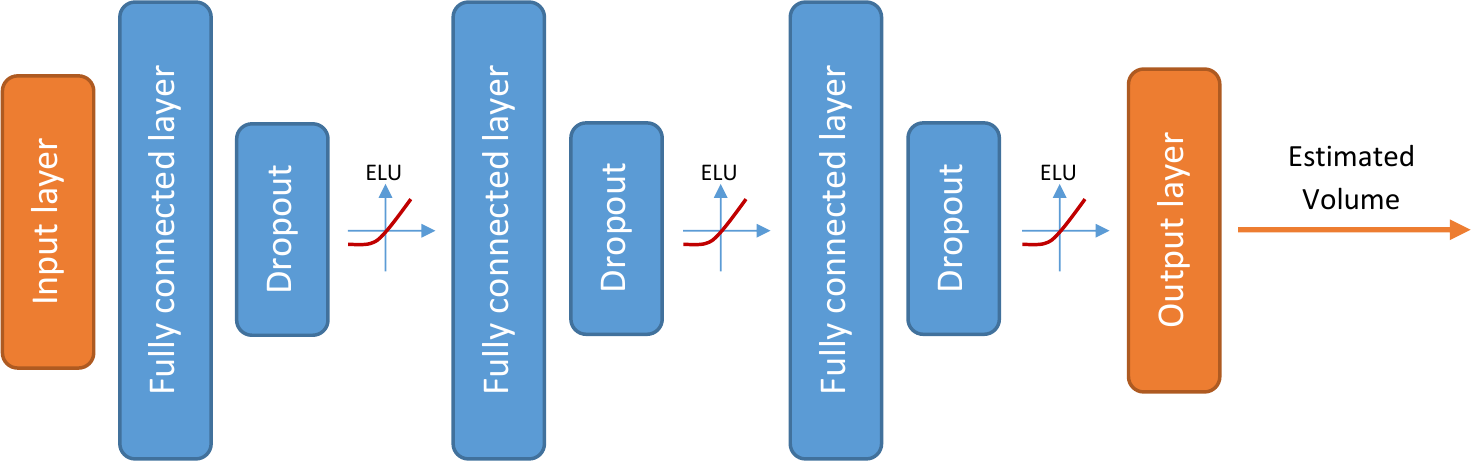}
	\caption{A fully-connected ANN with ELU activation is employed to estimate traffic volumes.}
	\label{Fig_fully_connected}
\end{figure}

The ANN's hyperparameters (i.e., number of layers and number of neurons in the hidden layers) were selected based on preliminary experiments. The final model configuration is comprised of an input layer, three hidden layers with 256 neurons each, and an output layer (Figure \ref{Fig_fully_connected}). The model was trained and evaluated as described in the following subsection.

\subsection{Model evaluation}
The previously-described ANN model is trained on data points from 44 ATR stations, and then tested on data from the remaining ATR station. Model estimates are then compared against the actual (ATR) volumes at this location to evaluate model performance. The outlined procedure is repeated 45 times (holding out each station once) and model accuracy for the 45 test locations (i.e., 90 carriageways) is summarized using four measures discussed below.

\begin{remark}	
In the outlined procedure the ANN model is always evaluated on data points from a location that was NOT considered during training, which corresponds to the real-world application of the proposed model. Iterating through the procedure allows for the evaluation of the model in different scenarios and it also enables fair assessment of its performance on all 90 carriageways.
\end{remark}

\subsubsection{Coefficient of determination ($R^2$)}
The $R^2$ metric indicates the proportion of traffic volume variance that is explained by the ANN model. This measure is computed as
\begin{align}
R^2=1-\frac{\sum^n_{j=1}{{\left(y_j-\hat{y}_j\right)}^2}}{\sum^n_{j=1}{{\left(y_j-\overline{y}\right)}^2}},
\end{align}
where $y_j$ denotes an actual (ATR) volume, $\hat{y}_j$ is an ANN-based volume estimate, and $\bar{y}$ represents the sample average. Values closer to 1 indicate better model performance.

\subsubsection{Mean absolute percentage error ($\textit{MAPE}$)}
$\textit{MAPE}$ is a measure of the relative accuracy of the model. It is defined as
\begin{align}
\textit{MAPE}=\left(\frac{1}{n}\sum^n_{j=1}{\left|\frac{{\hat{y}}_j-y_j}{y_j}\right|}\right)\times 100,
\end{align}
and it varies between 0 and 100 percent. Smaller values indicate better model performance. 

\subsubsection{Error-to-theoretical-capacity ratio ($\textit{ETCR}$)}
$\textit{ETCR}$ is a measure of a model's accuracy relative to the maximum sustainable traffic volume at a given location under prevailing road and traffic conditions (i.e., capacity). It was identified in an earlier survey \citep{ITSpaper2017} as a desirable accuracy measure from a practitioner's perspective. The same survey suggested that the desirable threshold for this error measure is 10\% for most planning and operations applications. It is defined as
\begin{align}
\textit{ETCR}=\left(\frac{1}{n}\sum^n_{j=1}{\left|\frac{{\hat{y}}_j-y_j}{c}\right|}\right) \times 100,
\end{align}
where $c$ denotes default/theoretical capacity estimates for the relevant facility from the Highway Capacity Manual \citep{manual2016}. Note that $c$ represents a function of free-flow speed at a given location and is expressed in passenger cars per hour per lane (Table \ref{TableHCM}). The free-flow speeds needed to compute $c$ were estimated based on vehicle probe data and speed limits. Finally, $\textit{ETCR}$ varies between 0 and 100 percent, and smaller values indicate better model performance. 

\begin{table}
	\centering
	\caption{Freeway and Multilane Highway Segment Capacity Conditions \citep{manual2016}}\label{TableHCM}
	\bgroup
	\def\arraystretch{1.25}% 
	{\footnotesize
		\begin{tabular}{ccc} \hline 
			Free-flow speed & Freeway Capacity & Multilane Highway Capacity\\ 
			(mi/h) & (pc/h/ln) & (pc/h/ln) \\ 
			\hline 
			75 & 2,400 & Not applicable \\  
			70 & 2,400 & 2,300 \\  
			65 & 2,350 & 2,300 \\  
			60 & 2,300 & 2,200 \\  
			55 & 2,250 & 2,100 \\ 
			50 & Not applicable & 2,000 \\  
			45 & Not applicable & 1,900 \\ \hline 
		\end{tabular}
	}
	\egroup
\end{table}

\subsubsection{Error-to-maximum-flow ratio ($\textit{EMFR}$)}
$\textit{EMFR}$ is similar to $\textit{ETCR}$; however, the difference is that errors are measured relative to the maximum observed volume at each location. It is defined as
\begin{align}
\textit{EMFR}= \left(\frac{1}{n}\sum^n_{j=1}{\left|\frac{{\hat{y}}_j-y_j}{y_{max}}\right|}\right)\times 100,
\end{align}
where $y_{max}$ is the maximum recorded volume at the observed location. Again, this measure varies between 0 and 100 percent, and smaller values indicate better model performance.

\subsection{GPU Implementation}
Applying the leave-one-ATR-out cross-validation approach ensures extensive evaluation of the model; however, it also implies extensive computation times to train forty five deep networks. To make the analysis computationally tractable, two solutions were employed. First, the ELU activation layer and Adam algorithm were used to increase efficiency of the training procedure itself. Additionally, computations were performed with a graphical processing (GPU) unit, which significantly speeds up computations that can be carried out in parallel (e.g., matrix multiplication). The computations were performed on Nvidia GeForce 1080, with the use of CUDA, CUDNN and TensorFlow libraries.

\section{Results}
Performance of the proposed ANN model is first compared against other machine learning models. Next, the ANN model is contrasted against the profiling method, which represents the current state-of-the-practice. Subsequently, the effect of vehicle probe volumes and their penetration rates on ANN model performance is assessed, which provides important managerial insights for transportation agencies interested in acquiring vehicle probe data for volume estimation purposes.

\subsection{The ANN vs. other machine learning models}
Here we compare the proposed ANN against other widely used models to assess its performance. Specifically, we compare it against the linear regression (LR), $k$-nearest neighbors ($k$-NN), support vector machines (SVM) with linear kernel, random forest (RF), and an ANN model without batch normalization (BN). Table \ref{TableMLcomparison} summarizes performance of all the models across 45 test locations (i.e., 90 carriageways). In terms of median $\textit{MAPE}$, the ANN model without BN performs the best; however, it also includes significant outlier locations with the maximum $\textit{MAPE}$ over 250\%. Hence, we still give preference to the proposed model with BN, even though its median performance is somewhat weaker (i.e., 24.59\% vs. 23.27\%). In addition, we consider two ensembles that average predictions from (a) the two ANN models and (b) the two ANNs and RF models. They show better median accuracy, but still suffer from poor worst-case performance and increased computation time needed to train more than one model on a relatively large data set. Hence, the ANN (with BN) was selected as the most reasonable choice when weighing in both its overall performance and more straightforward implementation, which are both very relevant for real-world deployment and application.

\begin{table}
	\centering
	\caption{Comparison of the proposed ANN model against other methods via $\textit{MAPE}$}
	\label{TableMLcomparison}
	\bgroup
	\def\arraystretch{1.25}% 
	{\footnotesize	
		\begin{tabular}{lllllllllll}\hline
			Measure & LR     & 1-NN   & 4-NN   & 16-NN  & SVM    & RF     & ANN    & \begin{tabular}[c]{@{}l@{}}ANN \\ (no BN)\end{tabular} & \begin{tabular}[c]{@{}l@{}}ensamble \\ ANNs\end{tabular} & \begin{tabular}[c]{@{}l@{}}ensamble \\ ANNs and RF\end{tabular} \\ \hline
			Min    & 15.61  & 6.02   & 9.03   & 12.94  & 16.86  & 10.52  & 11.05  & 10.60                                                  & 10.53                                                    & 9.88                                                            \\
			25th   & 26.04  & 31.18  & 25.14  & 25.25  & 25.00  & 19.05  & 17.40  & 18.27                                                  & 17.65                                                    & 16.74                                                           \\
			Median & 39.10  & 40.09  & 36.58  & 36.58  & 33.79  & 27.19  & 24.59  & 23.27                                                  & 22.27                                                    & 22.86                                                           \\
			75th   & 50.29  & 57.95  & 53.45  & 55.52  & 41.36  & 32.59  & 29.58  & 29.73                                                  & 29.13                                                    & 29.65                                                           \\
			Max    & 345.89 & 325.89 & 306.81 & 337.55 & 311.60 & 131.66 & 114.13 & 255.33                                                 & 179.36                                                   & 161.04 \\ \hline                                                         
		\end{tabular}
	}
	\egroup
\end{table}

\subsection{The ANN vs. profiling method}
\begin{figure}
	\centering
	\begin{subfigure}{0.48\textwidth}
		\centering
		\includegraphics[height=57mm]{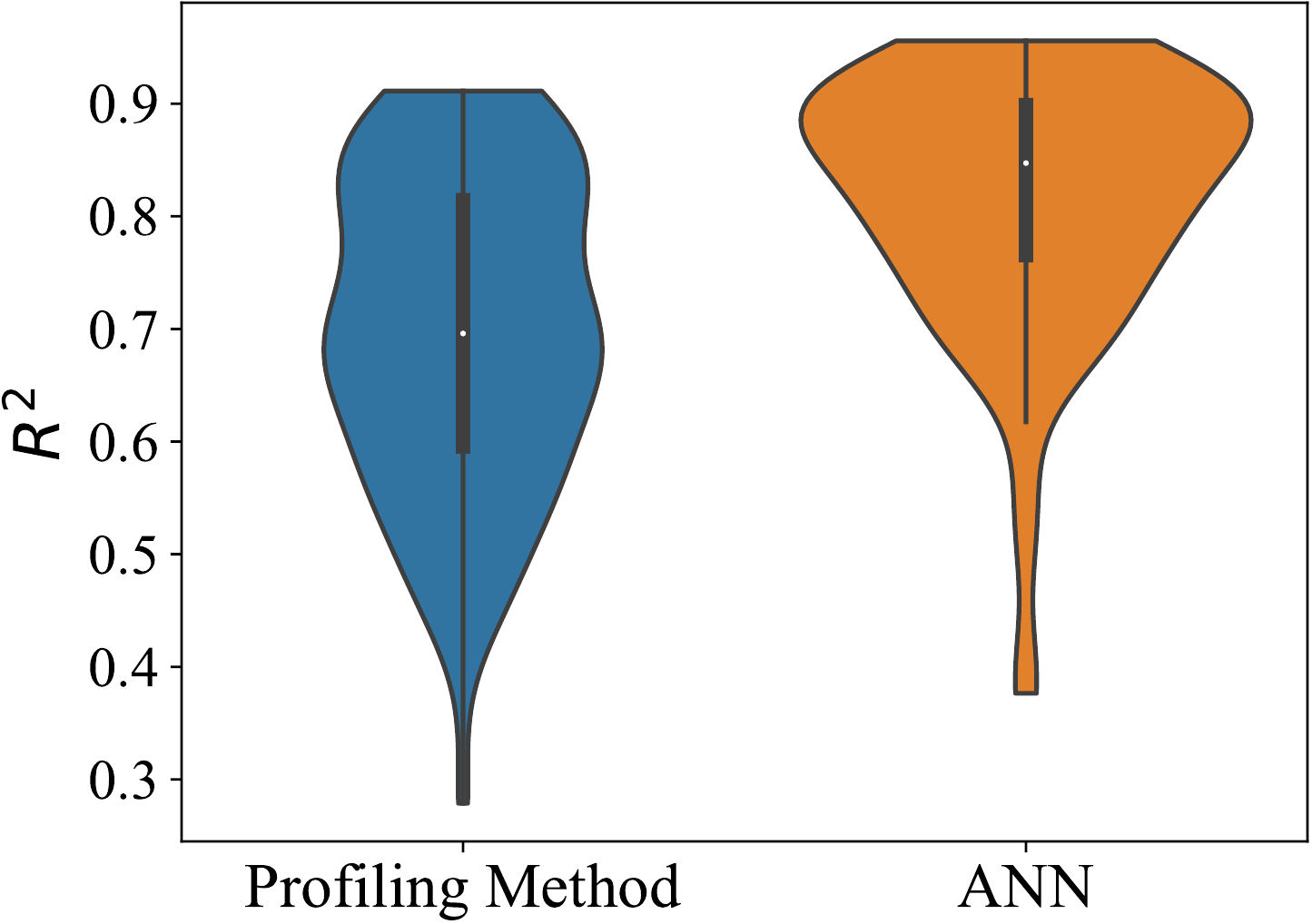}
		\caption{Comparison for $R^2$ (higher is better).}
	\end{subfigure} \hfill
	\begin{subfigure}{0.48\textwidth}
		\centering
		\includegraphics[height=57mm]{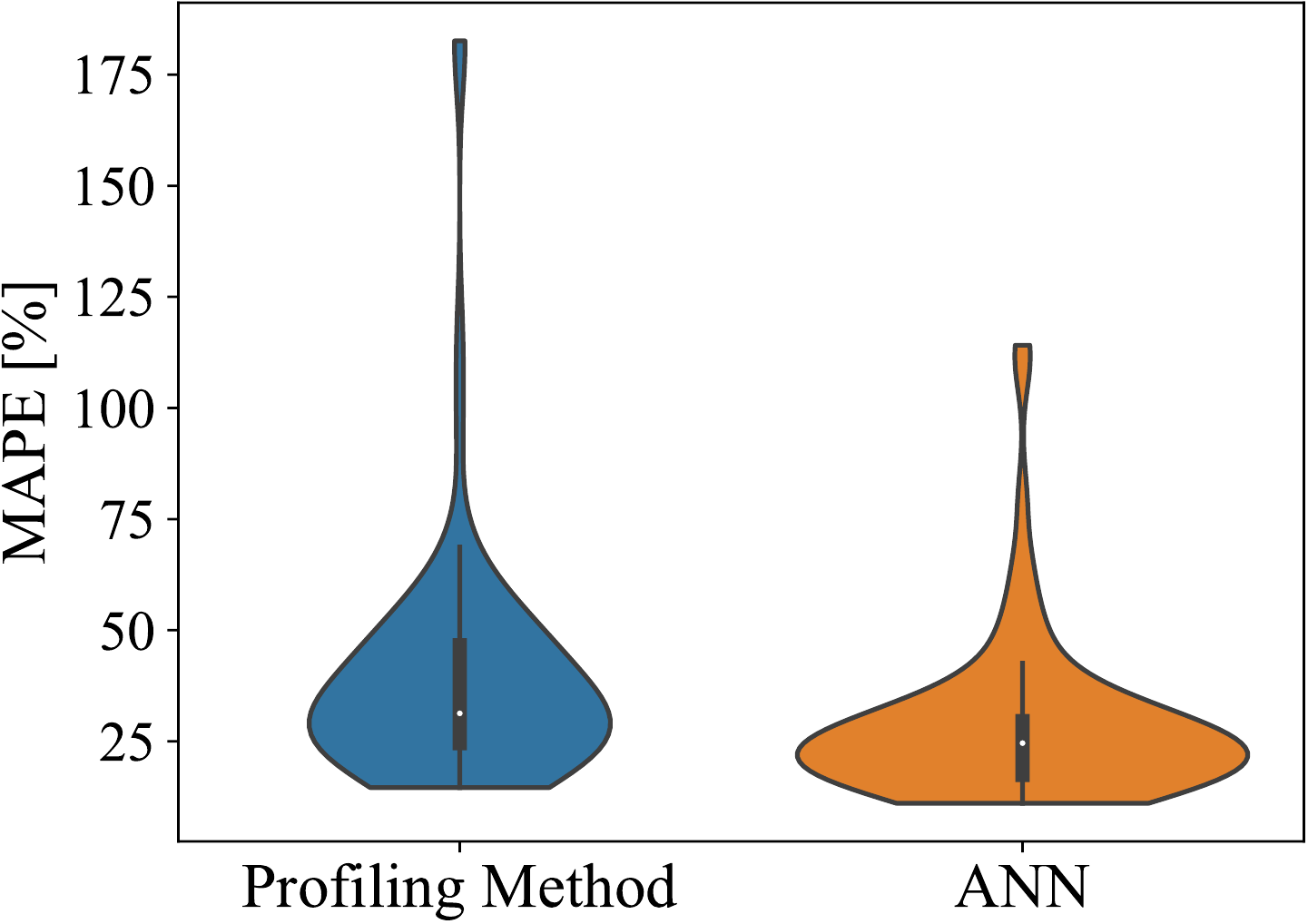}
		\caption{Comparison for $\textit{MAPE}$ (smaller is better).}
	\end{subfigure}
	
	\vspace{15pt}
	
	\begin{subfigure}{0.48\textwidth}
		\centering
		\includegraphics[height=57mm]{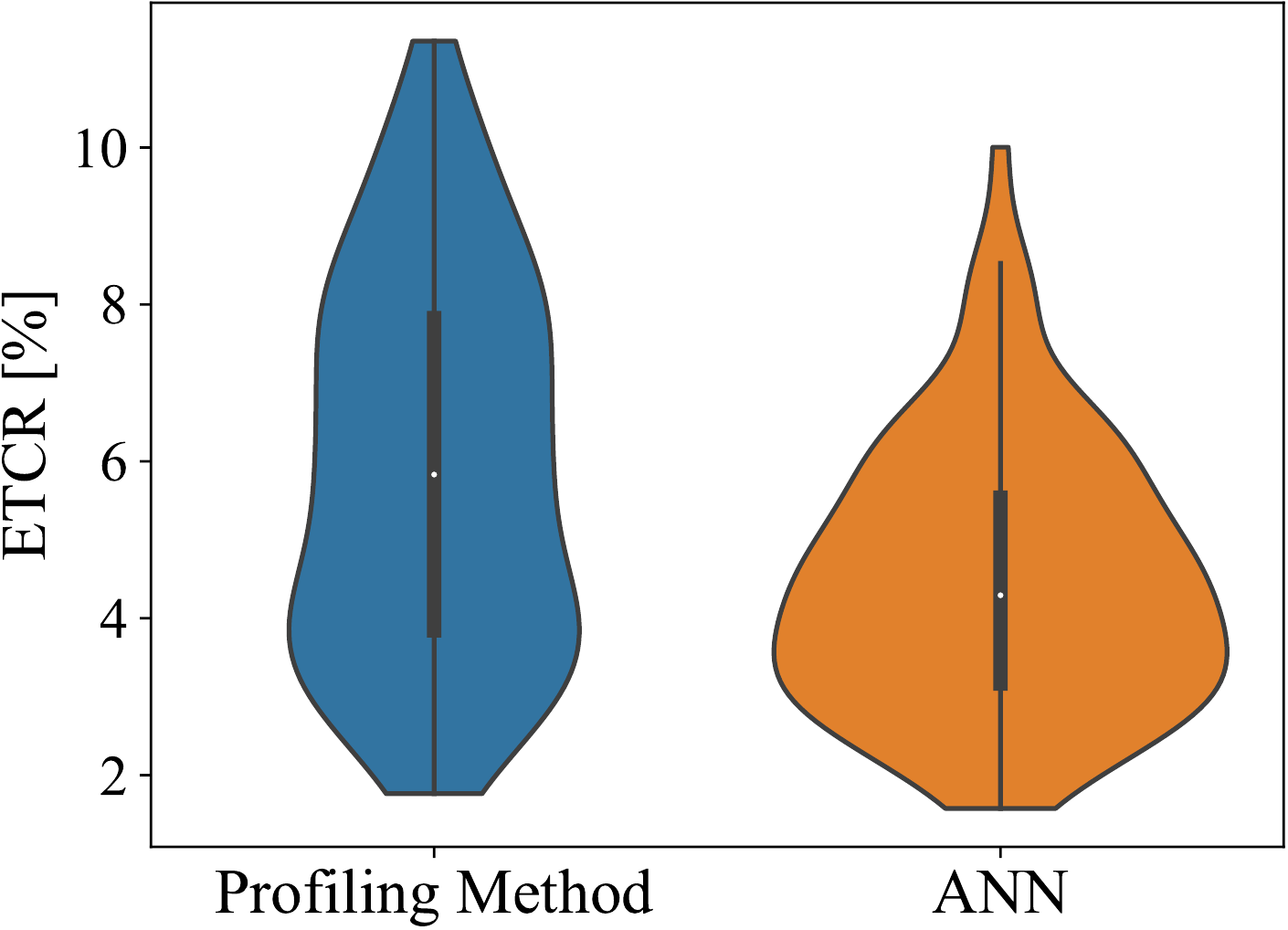}
		\caption{Comparison for $\textit{ETCR}$ (smaller is better).}
	\end{subfigure} \hfill
	\begin{subfigure}{0.48\textwidth}
		\centering
		\includegraphics[height=57mm]{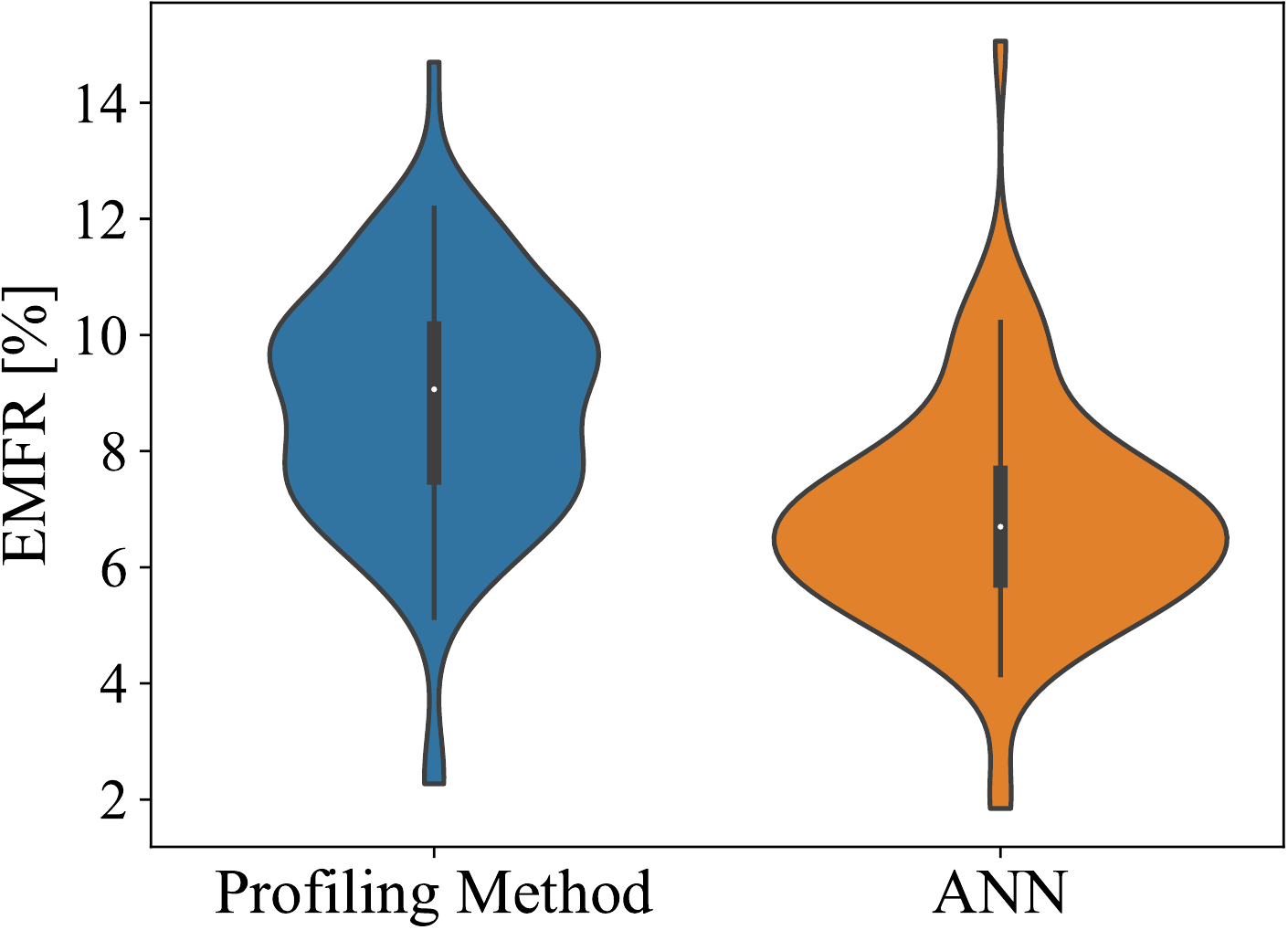}
		\caption{Comparison for $\textit{EMFR}$ (smaller is better).}
	\end{subfigure}

	\caption{Violin plots comparing overall performance of the two models at 90 carriageways used for testing.}\label{ViolinPlotsForTable2}
\end{figure}

\begin{figure}
	\centering
	\includegraphics[height=75mm]{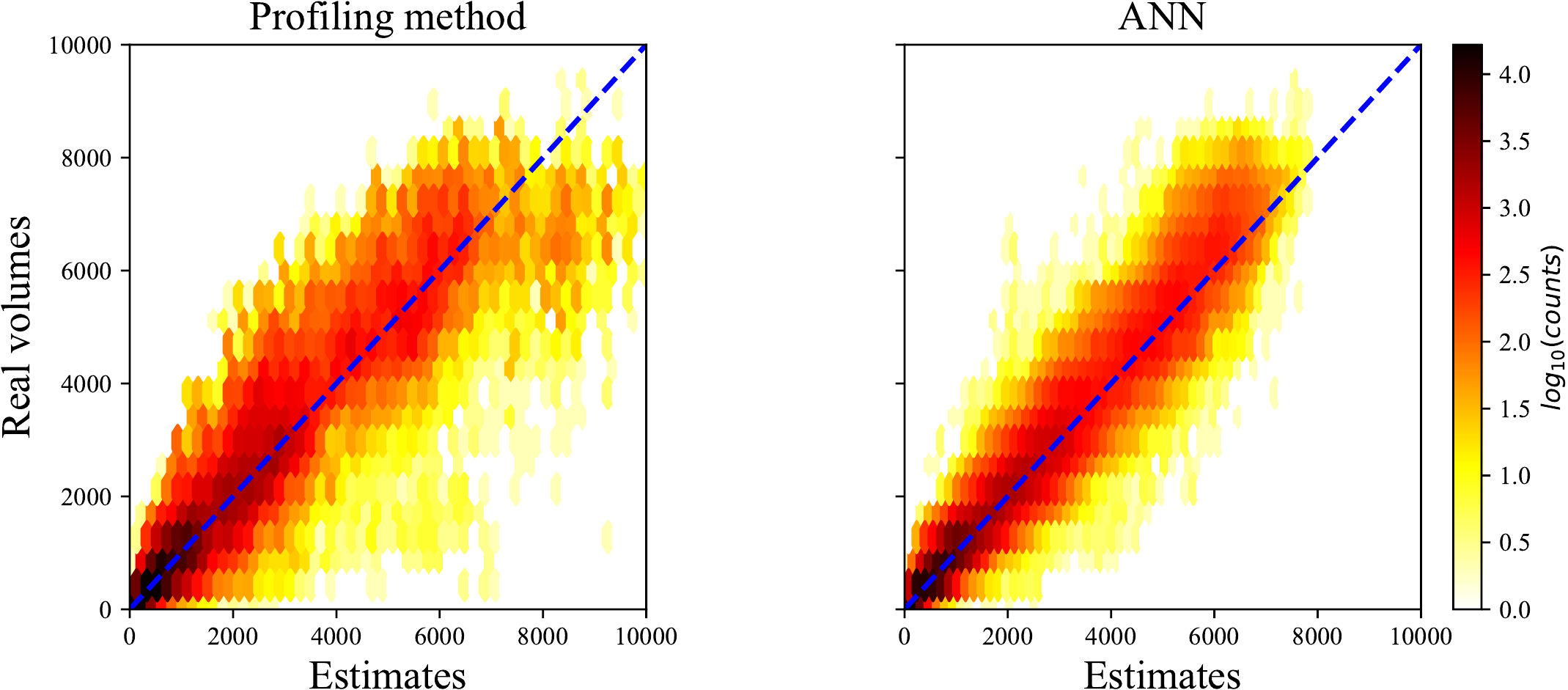}
	\caption{Heat maps comparing both model performance on all data points used for testing.}\label{ScatterPlanAndHeatMap}
\end{figure}

The aggregate performance of the two approaches for the 45 test locations (i.e., 90 carriageways) is summarized in Table \ref{TablePerformance}. Comparison of the four measures' median values indicates that the ANN outperforms the profiling method by 22\% ($R^2$), 21\% ($\textit{MAPE}$), 26\% ($\textit{ETCR}$), and 26\% ($\textit{EMFR}$), which corresponds to \ul{an average improvement of about 24\% across the four measures}. In addition, the Wilcoxon signed rank test \citep{gibbons2011nonparametric} rejects the null hypothesis that the median difference for each of the the four measures is zero at the default 5\% significance level. An extended comparison of the two approaches is shown via violin plots (Figure \ref{ViolinPlotsForTable2}), which visualize the distribution of the four measures over all 90 test carriageways and reiterate that the ANN model shows superior performance. This can also be seen in Figure \ref{ScatterPlanAndHeatMap}, which indicates that ANN-based estimates are more centered along the 45$^{\circ}$ line (i.e., where estimates are equal to actual (ATR) volumes). 

\begin{remark}	
	The proposed ANN model outperforms the widely-used profiling method by about 24\% across the four performance measures.
\end{remark}

\begin{table}
	\centering
	\caption{The overall performance of the ANN model and the widely-used profiling method}
	\label{TablePerformance}
	\bgroup
	\def\arraystretch{1.25}% 
	{\footnotesize	
		\begin{tabular}{llllllllll}\hline 			
			\multirow{2}{*}{Measure} & \multicolumn{4}{c}{Profiling Method} && \multicolumn{4}{c}{ANN} \\ \cline{2-5} \cline{7-10}
			& $R^2$ & $\textit{MAPE}$ & $\textit{ETCR}$ & $\textit{EMFR}$ && $R^2$ & $\textit{MAPE}$ & $\textit{ETCR}$ & $\textit{EMFR}$ \\ \hline		
			Minimum                  & 0.28 & 14.60\%  & 1.76\%  & 2.27\%  &  & 0.38 & 11.05\%  & 1.57\%  & 1.84\%  \\
			25th percentile          & 0.60 & 24.56\%  & 3.85\%  & 7.54\%  &  & 0.77 & 17.40\%  & 3.17\%  & 5.77\%  \\
			50th percentile (median) & 0.70 & 31.29\%  & 5.83\%  & 9.07\%  &  & 0.85 & 24.59\%  & 4.29\%  & 6.70\%  \\
			75th percentile          & 0.81 & 46.59\%  & 7.82\%  & 10.12\% &  & 0.90 & 29.58\%  & 5.54\%  & 7.63\%  \\
			Maximum                  & 0.91 & 182.61\% & 11.36\% & 14.70\% &  & 0.96 & 114.13\% & 10.00\% & 15.06\% \\ \hline			
			
		\end{tabular}
	}
	\egroup
\end{table}

\subsection{The effect of probe volume intensity}

\begin{figure}
	\centering
	\begin{subfigure}{0.48\textwidth}
		\centering
		\includegraphics[height=62mm]{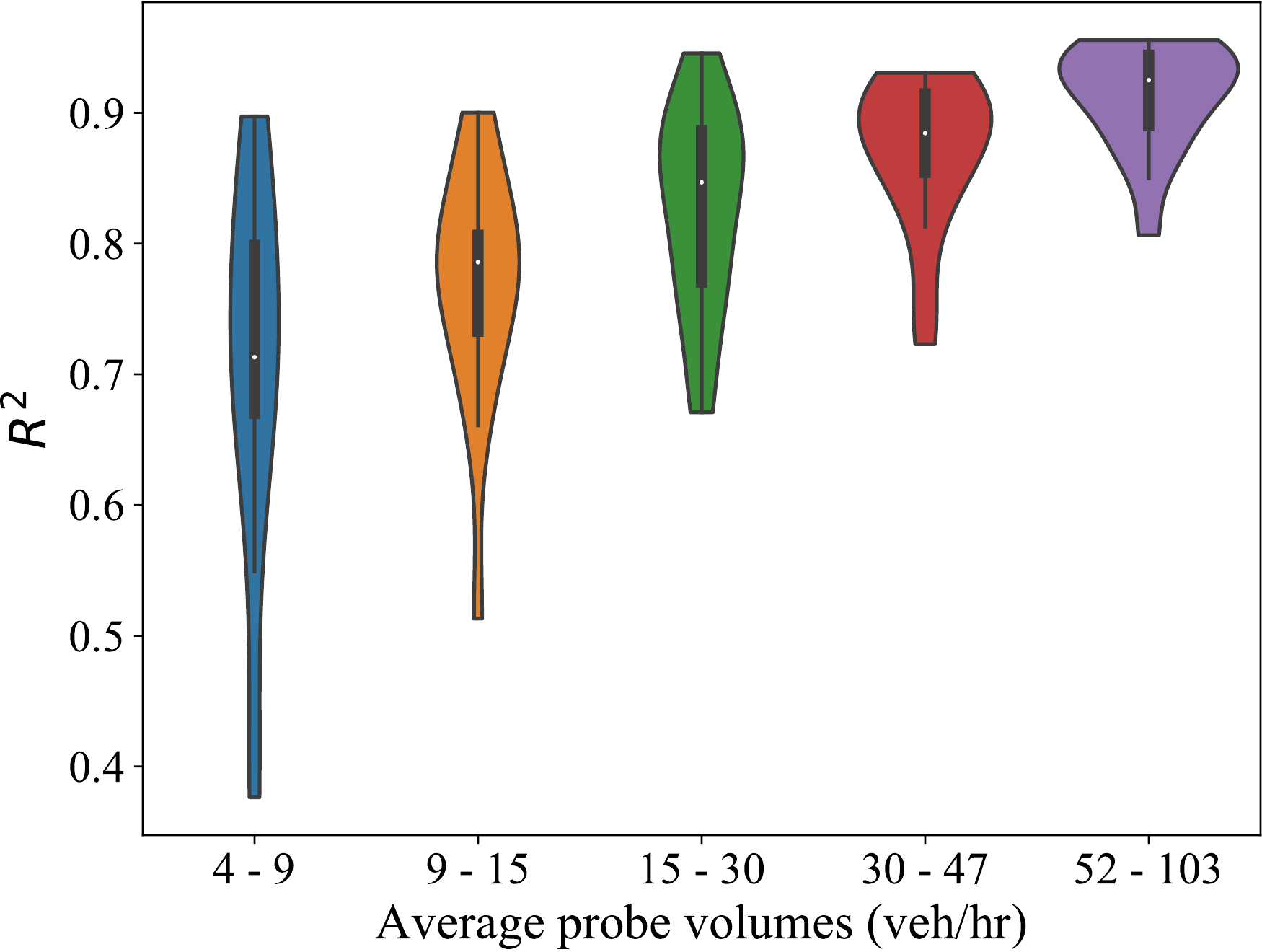}
		\caption{Comparison for $R^2$ (higher is better).}
	\end{subfigure}\hfill
	\begin{subfigure}{0.48\textwidth}
		\centering
		\includegraphics[height=62mm]{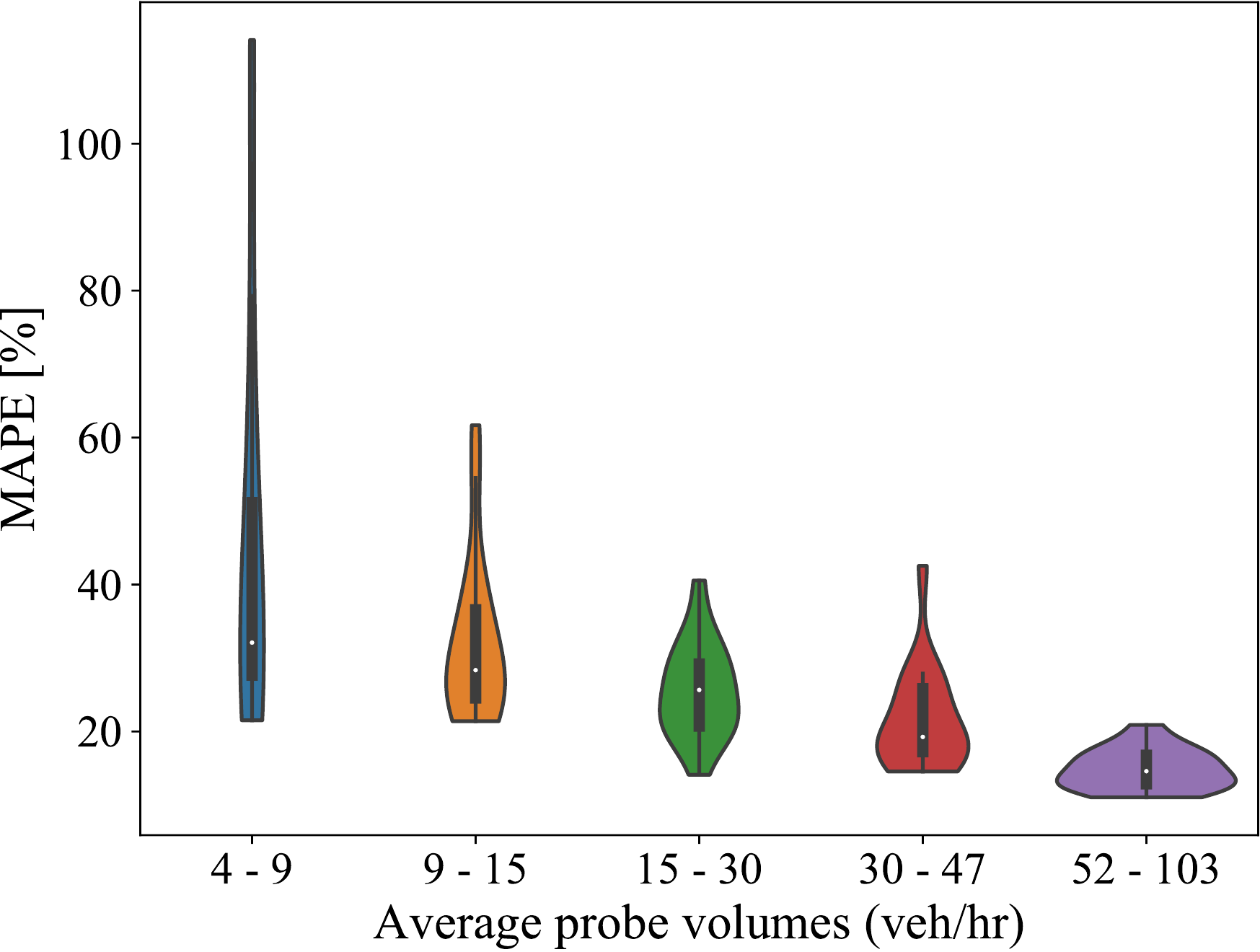}
		\caption{Comparison for $\textit{MAPE}$ (smaller is better).}
	\end{subfigure}
	
	\vspace{15pt}
	
	\begin{subfigure}{0.48\textwidth}
		\centering
		\includegraphics[height=62mm]{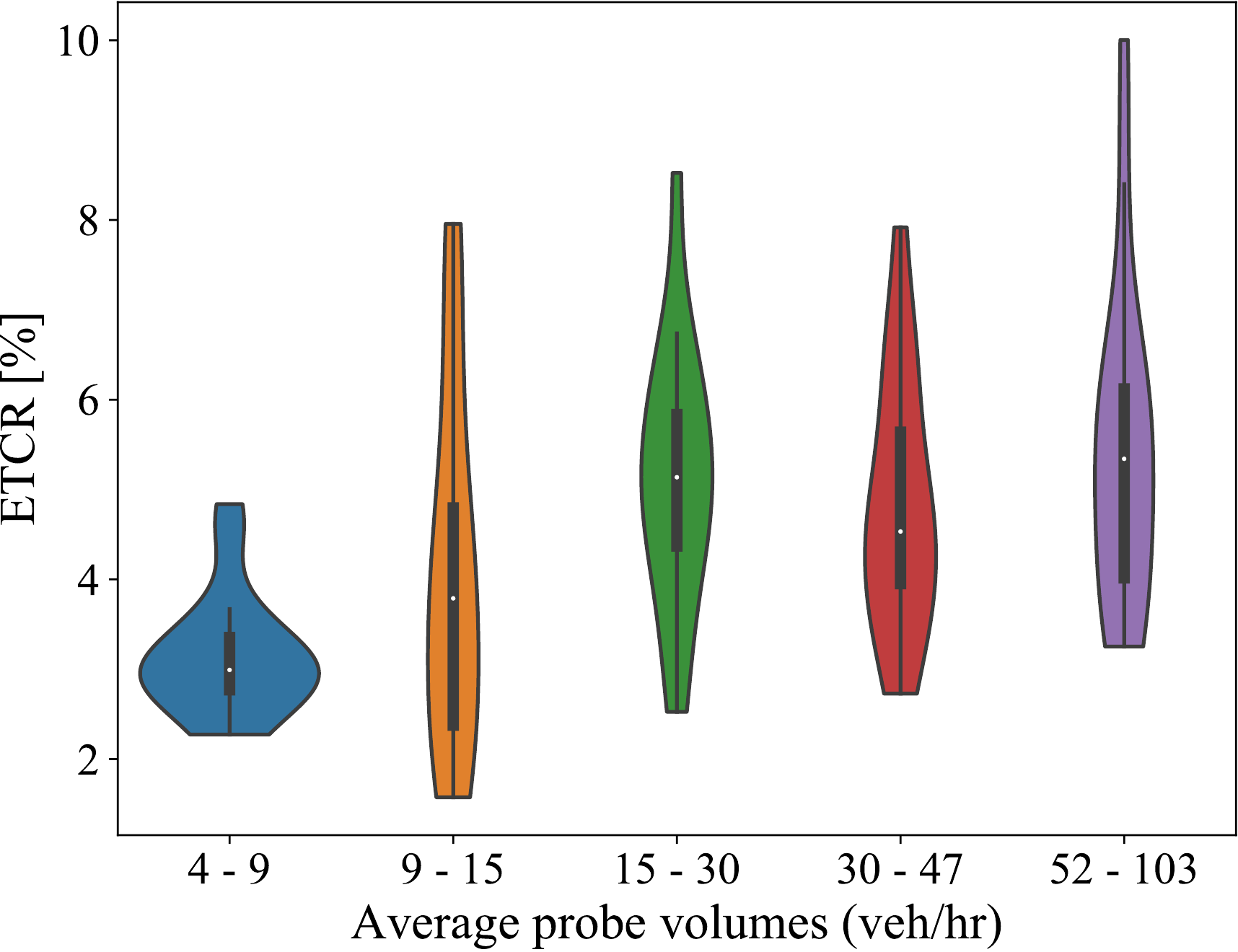}
		\caption{Comparison for $\textit{ETCR}$ (smaller is better).}\label{Figure4c}
	\end{subfigure}\hfill
	\begin{subfigure}{0.48\textwidth}
		\centering
		\includegraphics[height=62mm]{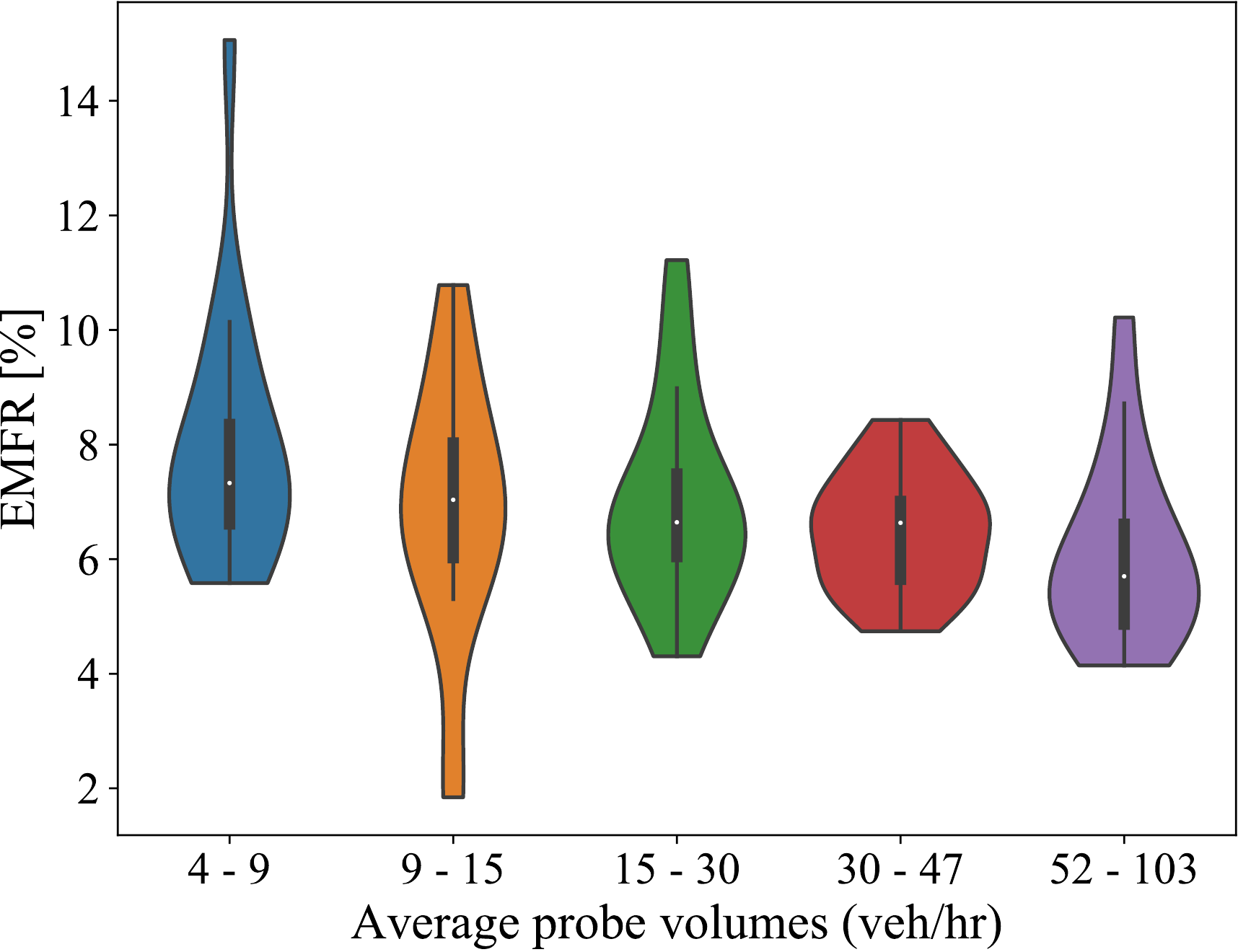}
		\caption{Comparison for $\textit{EMFR}$ (smaller is better).}\label{Figure4d}
	\end{subfigure}
	
	\caption{The violin plots showing distribution of ANN's performance measures at locations with different average volumes of probe vehicles.}\label{ViolinPlotsForTable3}
\end{figure}

The performance of the ANN model is then evaluated for different probe vehicle intensities, and compared against the profiling method (Table \ref{TableProbePerformance}). Specifically, the 90 test carriageways are divided equally into five groups based on their average vehicle probe volumes, and the four performance measures are computed for both methods (i.e., ANN and profiling) while considering each group separately. Results show that the ANN outperforms the profiling method for each group and all four measures. This suggests that probe data is an important model input, and the ANN's accuracy depends on having sufficient number of probe vehicles. Moreover, it is noticeable that the $R^2$ and $\textit{MAPE}$ for both methods improve as the average vehicle probe volumes are increased. The aforementioned pattern is particularly intuitive in the case of the ANN, and can also be observed in Figure \ref{ViolinPlotsForTable3} which shows the distribution of the four measures at locations with different vehicle probe volumes. The violin plots comparing $R^2$ and $\textit{MAPE}$ clearly show (in terms of the median and the shape of the entire distribution) that the ANN model provides better volume estimates at locations with higher average vehicle probe volumes. On the other hand, $\textit{ETCR}$ and $\textit{EMFR}$ are relatively insensitive to the changes in vehicle probe volumes, which is unsurprising because they are computed using maximum theoretical/observed traffic volumes -- quantities which are strongly correlated with probe volumes (provided stable penetration rates).

\begin{remark}	
	The ANN model outperforms the profiling method at locations with different average vehicle probe volumes. At locations where the average number of probes is between 30 and 47 vehicles/hr, the ANN achieves precision measured with the median $\textit{MAPE}$ of about 21\%. The median $\textit{MAPE}$ drops to about 15\% at locations where the average number of observed probes is between 52 and 103 vehicles/hr.
\end{remark}
 
\begin{table}
	\centering
	\caption{The effect of average probe volume intensity on median measures}
	\label{TableProbePerformance}
	\bgroup
	\def\arraystretch{1.25}% 
	{\footnotesize	
		\begin{tabular}{ccccccccccc}
			\hline 
			\multirow{2}{*}{Probes/hr} & \multicolumn{4}{c}{Profiling Method} && \multicolumn{4}{c}{ANN} \\ \cline{2-5} \cline{7-10}
			& $R^2$ & $\textit{MAPE}$ & $\textit{ETCR}$ & $\textit{EMFR}$ && $R^2$ & $\textit{MAPE}$ & $\textit{ETCR}$ & $\textit{EMFR}$ \\ \hline	 			
		{[}4, 9{]}       & 0.62 & 58.64\% & 3.47\% & 8.74\% &  & 0.71 & 46.03\% & 3.14\% & 7.94\% \\
		{[}9, 14.6{]}    & 0.66 & 47.38\% & 5.11\% & 8.46\% &  & 0.77 & 31.91\% & 4.08\% & 6.84\% \\
		{[}15.4, 30{]}   & 0.67 & 37.06\% & 6.93\% & 9.31\% &  & 0.83 & 25.54\% & 5.12\% & 6.99\% \\
		{[}30.1, 46.9{]} & 0.74 & 29.80\% & 6.46\% & 8.81\% &  & 0.87 & 21.45\% & 4.80\% & 6.47\% \\
		{[}52, 103.2{]}  & 0.80 & 21.08\% & 7.61\% & 8.71\% &  & 0.91 & 14.83\% & 5.37\% & 6.09\% \\\hline 
		\end{tabular}
	}
	\egroup
\end{table}

\subsection{The effect of probe vehicle penetration rates}

\begin{figure}
	\centering
	\begin{subfigure}{0.48\textwidth}
		\centering
		\includegraphics[height=62mm]{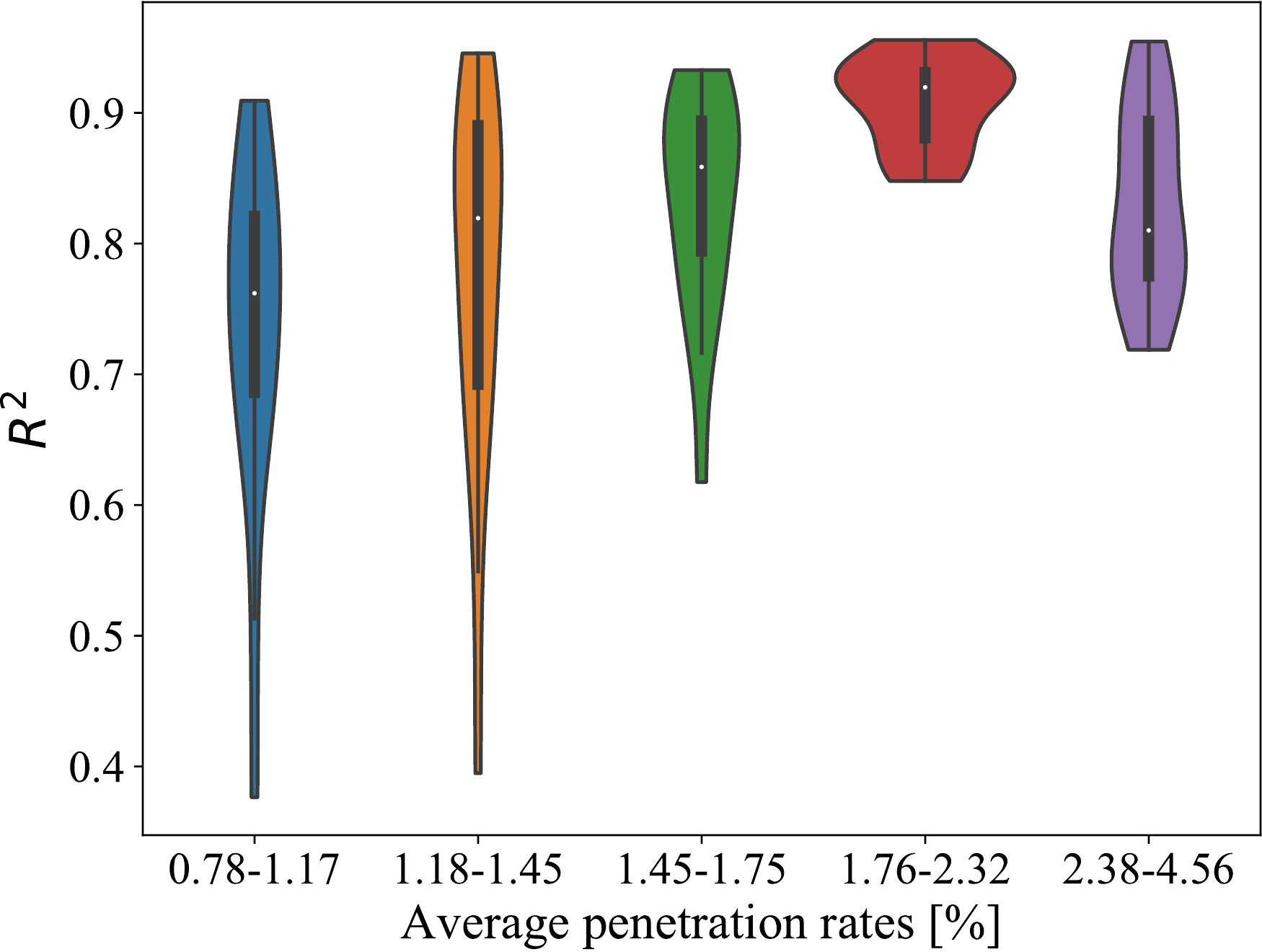}
		\caption{Comparison for $R^2$ (higher is better).}
	\end{subfigure}\hfill
	\begin{subfigure}{0.48\textwidth}
		\centering
		\includegraphics[height=62mm]{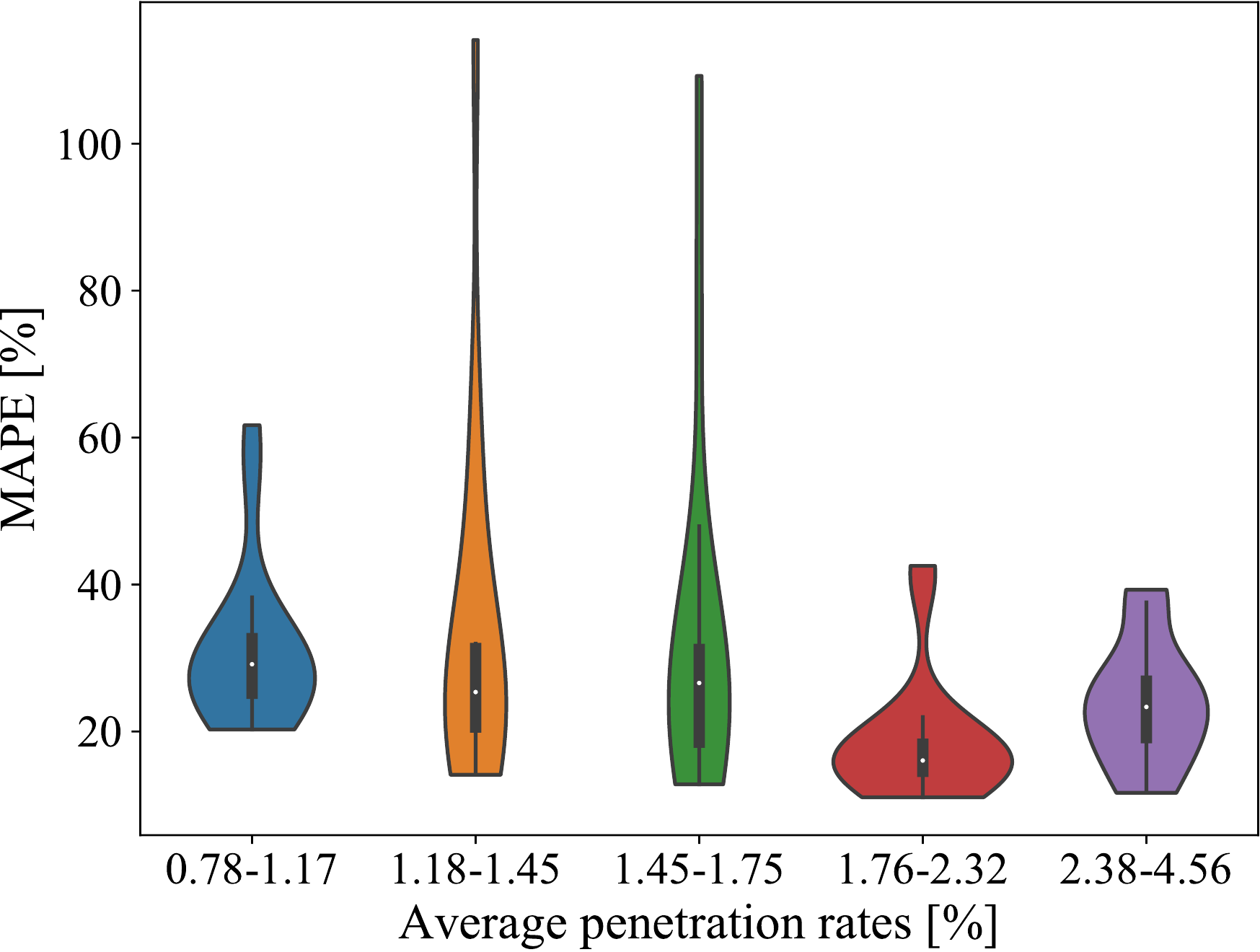}
		\caption{Comparison for $\textit{MAPE}$ (lower is better).}
	\end{subfigure}

	\vspace{15pt}
	
	\begin{subfigure}{0.48\textwidth}
		\centering
		\includegraphics[height=62mm]{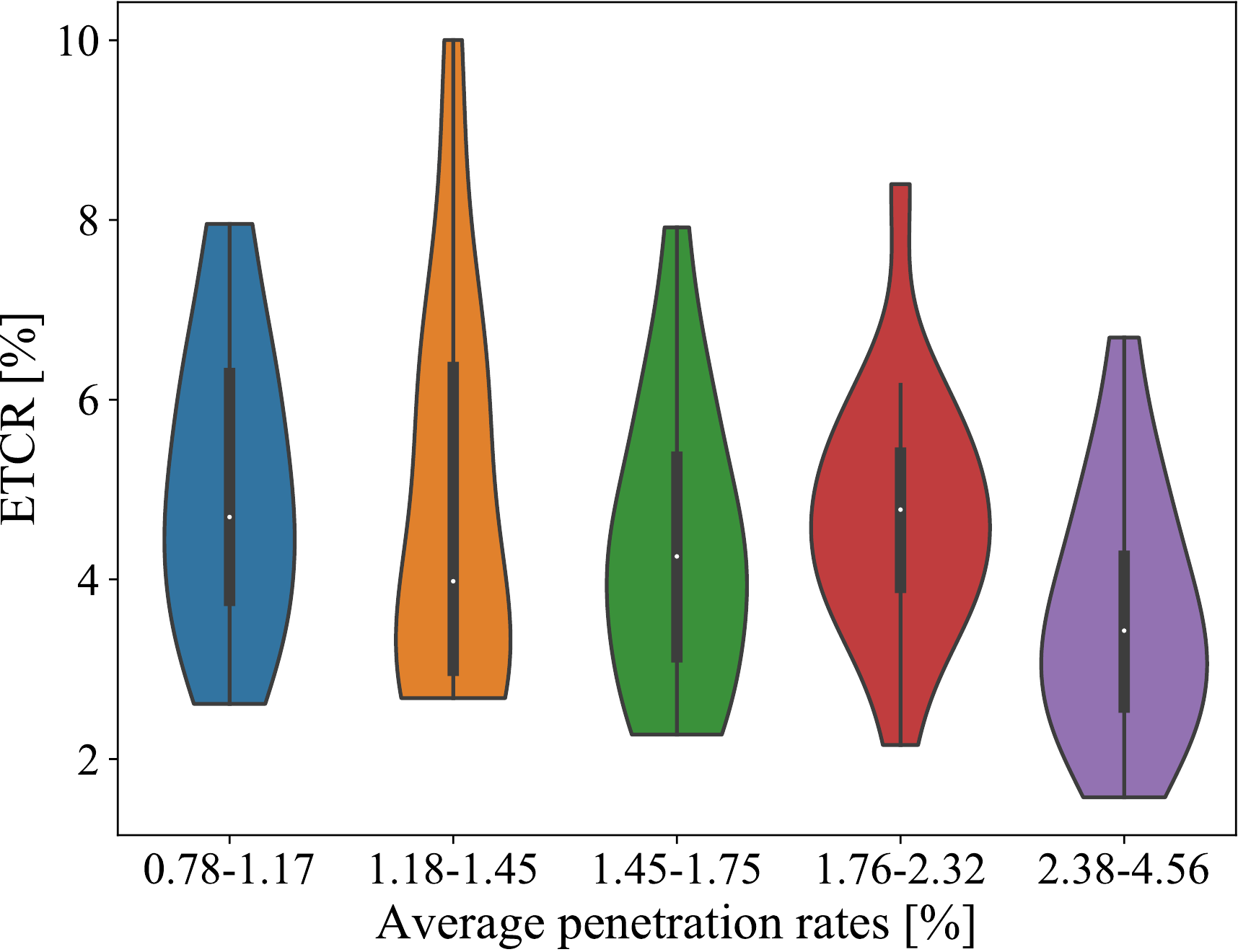}
		\caption{Comparison for $\textit{ETCR}$ (lower is better).}
	\end{subfigure}\hfill
	\begin{subfigure}{0.48\textwidth}
		\centering
		\includegraphics[height=62mm]{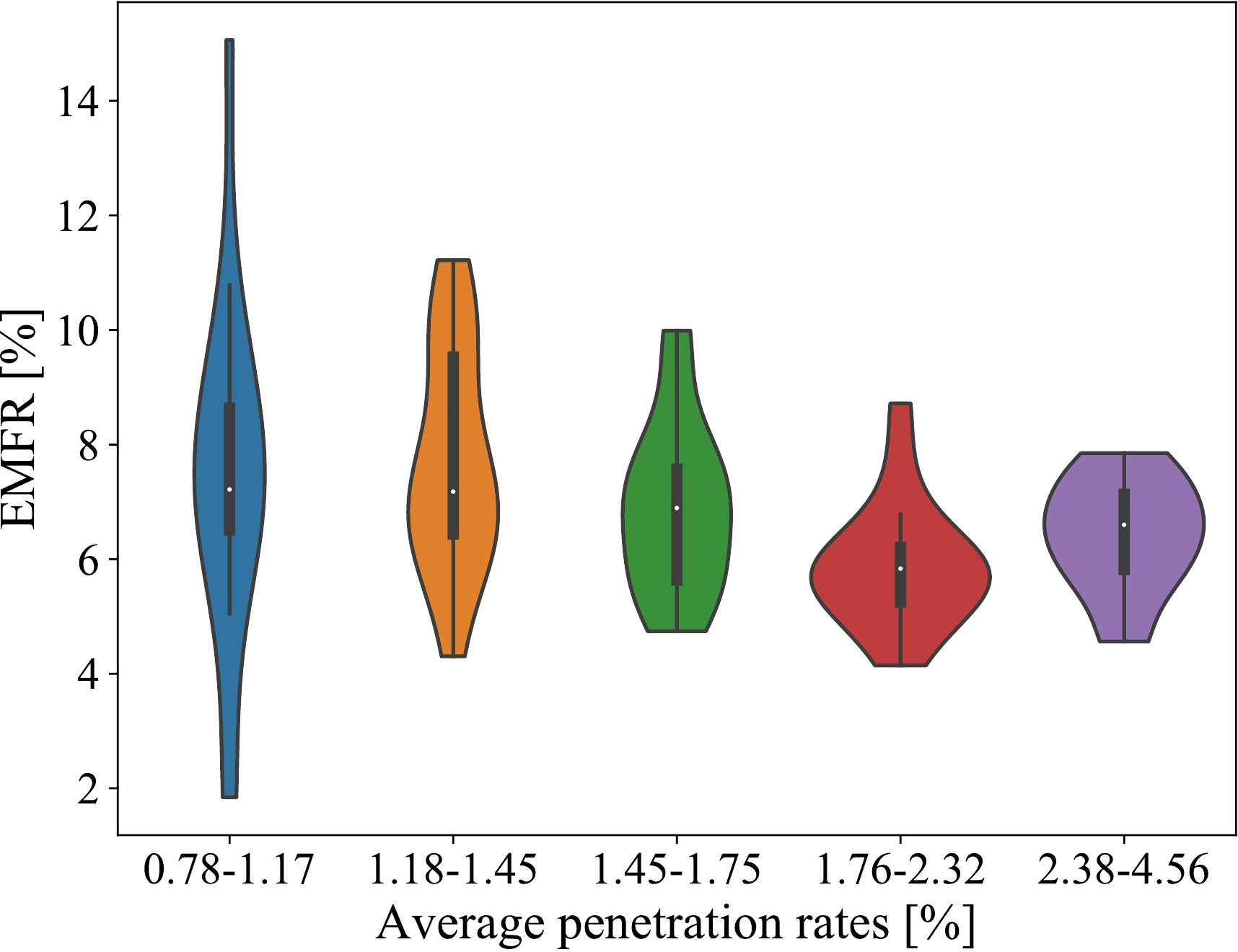}
		\caption{Comparison for $\textit{EMFR}$ (lower is better).}
	\end{subfigure}
	
	\caption{The violin plots showing distribution of ANN's performance measures at locations with different average penetration rates of probe vehicles.}\label{ViolinPlotsForTable4}
\end{figure}

A similar analysis is conducted to examine the performance of the ANN model given different penetration rates of probe vehicles (Table \ref{TablePenetrationPerformance}). Again, the 90 test carriageways are divided equally into five groups, but this time based on the average penetration rate of probe vehicles. The four performance measures are again computed for both methods (i.e., ANN and profiling) while considering each penetration rate group separately. Results show that the ANN outperforms the profiling method for each group and all measures, and that both models generally provide higher $R^2$ and $\textit{MAPE}$ values at locations with a greater penetration rate of probe vehicles. These intuitive conclusions are also noticeable in Figure \ref{ViolinPlotsForTable4}, which shows the distribution of the ANN's performance measures at locations with different penetration rates of probe vehicles. The violin plots show that the measures generally improve with higher penetration rates of probe vehicles. A noticeable exception is the last group for $R^2$, $\textit{MAPE}$ and $\textit{EMFR}$.

\begin{remark}	
	The ANN model outperforms the profiling method at locations with different average penetration rate of probe vehicles. At locations where the average penetration rate of vehicle probes is between 1.76\% and 4.56\%, the ANN achieves median $\textit{MAPE}$ of about 19-24\%.
\end{remark}

\begin{table}
	\centering
	\caption{The effect of average probe vehicle penetration rate on median measures}
	\label{TablePenetrationPerformance}
	\bgroup
	\def\arraystretch{1.25}% 
	{\footnotesize	
		\begin{tabular}{ccccccccccc}
			\hline 
			\multirow{2}{*}{Penetration} & \multicolumn{4}{c}{Profiling Method} && \multicolumn{4}{c}{ANN} \\ \cline{2-5} \cline{7-10} 
			& $R^2$ & $\textit{MAPE}$ & $\textit{ETCR}$ & $\textit{EMFR}$ && $R^2$ & $\textit{MAPE}$ & $\textit{ETCR}$ & $\textit{EMFR}$ \\ \hline	 			
			{[}0.78\%, 1.17\%{]} & 0.67 & 47.70\% & 6.12\% & 8.70\% &  & 0.74 & 31.03\% & 5.02\% & 7.43\% \\
			{[}1.18\%, 1.45\%{]} & 0.70 & 42.82\% & 5.49\% & 8.67\% &  & 0.78 & 33.50\% & 4.81\% & 7.70\% \\
			{[}1.46\%, 1.75\%{]} & 0.69 & 44.90\% & 6.04\% & 9.28\% &  & 0.83 & 32.54\% & 4.34\% & 6.90\% \\
			{[}1.76\%, 2.32\%{]} & 0.77 & 26.81\% & 7.00\% & 8.70\% &  & 0.91 & 18.86\% & 4.74\% & 5.89\% \\
			{[}2.38\%, 4.56\%{]} & 0.68 & 31.73\% & 4.95\% & 8.68\% &  & 0.83 & 23.81\% & 3.60\% & 6.41\% \\\hline 
		\end{tabular}
	}
	\egroup
\end{table}

%\subsection{Scaling results to entire Maryland road network}
%In the case of this paper, the proposed ANN model is evaluated on numerous data points from 45 locations (i.e., 90 carriageways) equipped with ATR stations, which provides both the ground-truth for training and testing the model. However, the main impetus for this approach is to train the model on the available ATR stations, and then use it to estimate historical hourly volumes throughout the state of Maryland. This would require mining the explanatory variables for each of the 11,000 traffic message channels in Maryland and simply feeding them to the ANN model. The former task would require significant processing effort to extract all of the data discussed in Section \ref{DataSection} (except for ATR counts which are clearly unavailable throughout the road network), while the latter task would be trivial and computationally efficient. Based on the encouraging results reported in this paper, the computationally-intense work may be carried out in collaboration with local agencies and their technical teams.

\section{Conclusions}
This paper addresses the problem of estimating historical hourly traffic volumes that transportation agencies need for planning and statewide performance measurement (e.g., user delay cost, energy efficiency). The proposed ANN-based approach significantly outperforms the profiling method often used to obtain hourly volume profiles from AADT data. The observed improvement of about 24\% is based on probe vehicle data that captures about 1.8\% of traffic, which indicates that application of vehicle probe data and machine learning is a promising approach towards improving the state-of-the-practice in estimating hourly traffic volumes. Moreover, results show that volumes can be estimated with $\textit{MAPE}$ of about 21\% at locations where average number of probes is between 30 and 47 vehicles/hr, and that $\textit{MAPE}$ reduces to about 15\% when observed number of probes is between 52 and 103 vehicles/hr. This provides a useful guideline for transportation agencies who wish to assess the value of probe vehicle data from different vendors. It is noteworthy that further GPS technology market penetration (especially cell phone location services) will increase traffic capture rates substantially, and thus improve volume estimation accuracy. Currently, a potential way to increase the capture rates and accuracy of the presented model is to merge vehicle probe and cell phone data from multiple providers. 

In the case of this paper, the proposed ANN model is evaluated on numerous data points from 45 locations (i.e., 90 carriageways) equipped with ATR stations, which provides both the ground-truth for training and testing the model. However, the main impetus for this approach is to train the model on the available ATR stations, and then use it to estimate historical hourly volumes throughout the state of Maryland. This would require mining the explanatory variables for each of the 11,000 traffic message channels in Maryland and simply feeding them to the ANN model. The former task would require significant processing effort to extract all of the data discussed in Section \ref{DataSection} (except for ATR counts which are clearly unavailable throughout the road network), while the latter task would be trivial and computationally efficient. Based on the encouraging results reported in this paper, the computationally-intensive work will be carried out as part of future work.

\section*{Acknowledgments}
The authors would like to thank the anonymous referees whose comments helped improve the paper. The authors are grateful to the I-95 Corridor Coalition for funding this work through the Volume and Turning Movement Project. Help from Subrat Mahapatra and the Maryland State Highway Administration who provided the vehicle probe data is also much appreciated. This support is gratefully acknowledged, but it implies no endorsement of the findings.

%\section*{References}
\bibliography{VolumeReferencesTRC}

\section*{Appendix}
Here we provide additional results that were used to (a) sanity check the proposed ANN model, (b) assess the effect of the dropout procedure.

\subsection{Checking for overfitting}

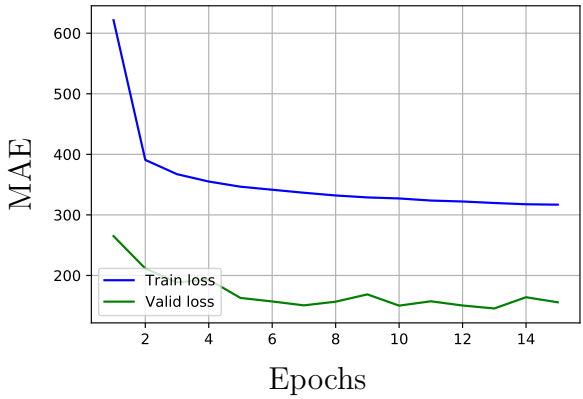
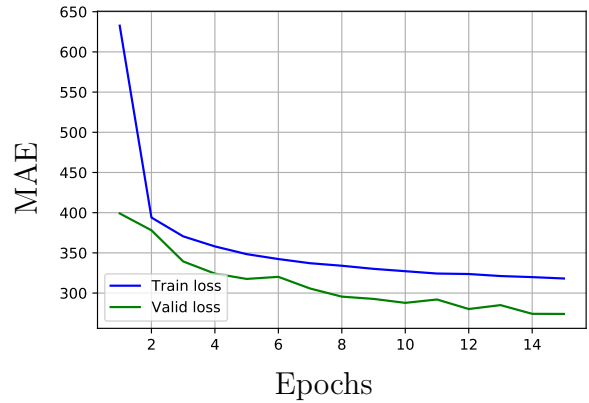
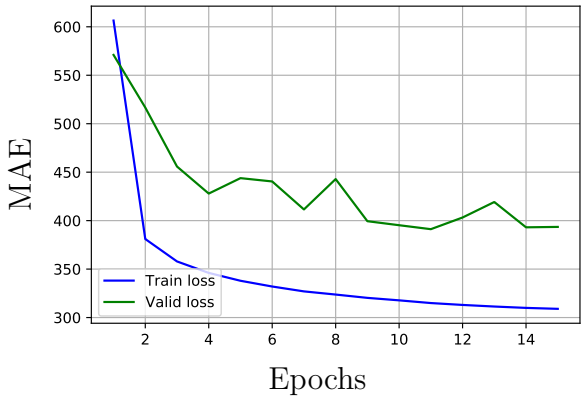
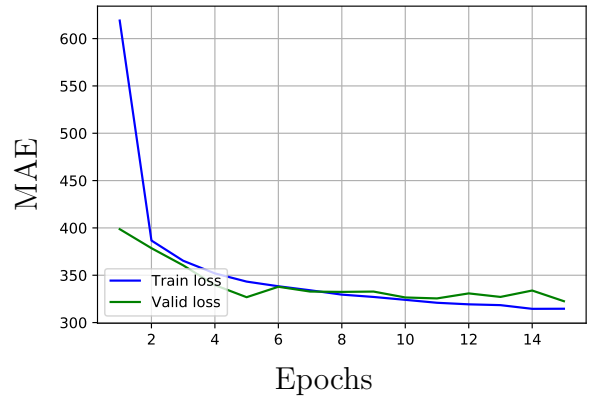
\begin{figure}
	\centering
	\begin{subfigure}{0.48\textwidth}
		\centering
		\begin{tikzpicture}
		\node (img)  {\includegraphics[scale=0.55]{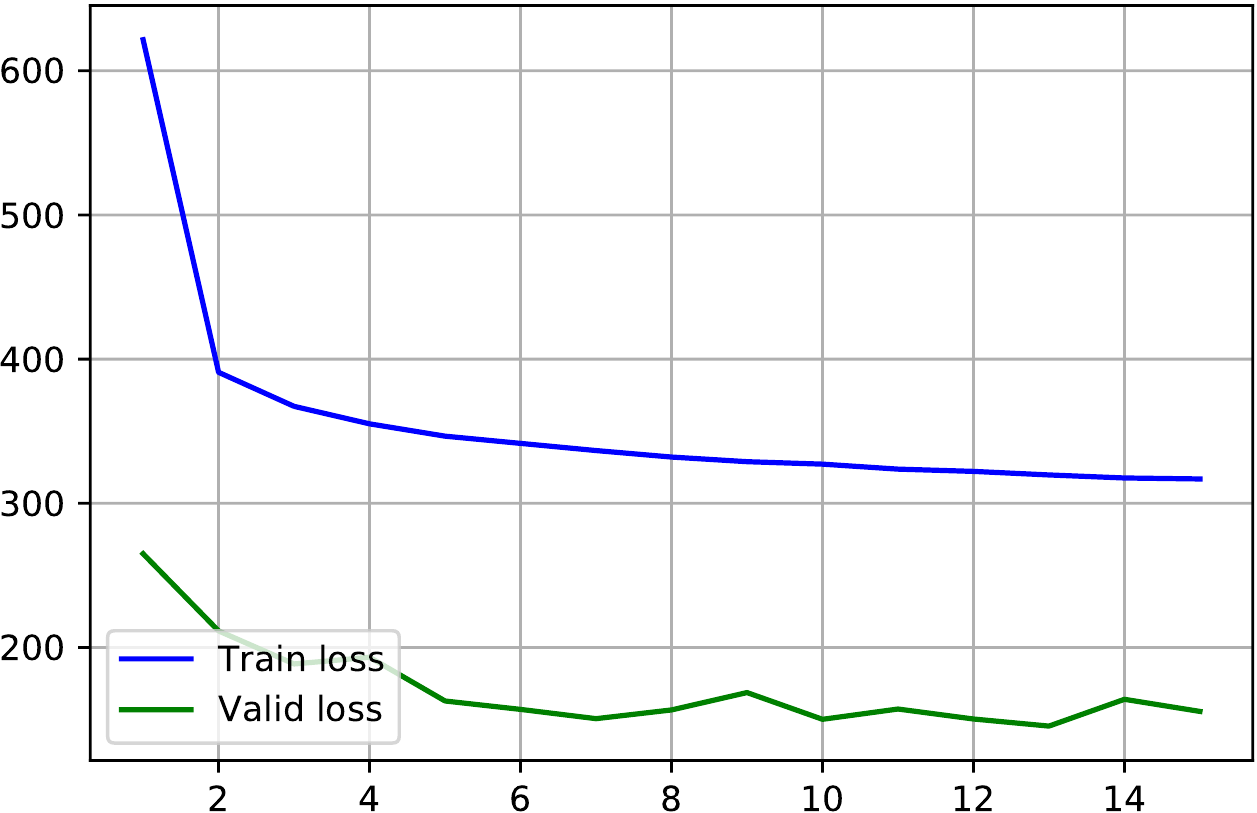}};
		\node[below=of img, node distance=0cm, yshift=1cm,font=\color{black}] {Epochs};
		\node[left=of img, node distance=0cm, rotate=90, anchor=center,yshift=-0.7cm,font=\color{black}] {MAE};
		\end{tikzpicture}
		\caption{Validation station 14.}\label{FigStation14}
	\end{subfigure}\quad
	\begin{subfigure}{0.48\textwidth}
		\centering
		\begin{tikzpicture}
		\node (img)  {\includegraphics[scale=0.55]{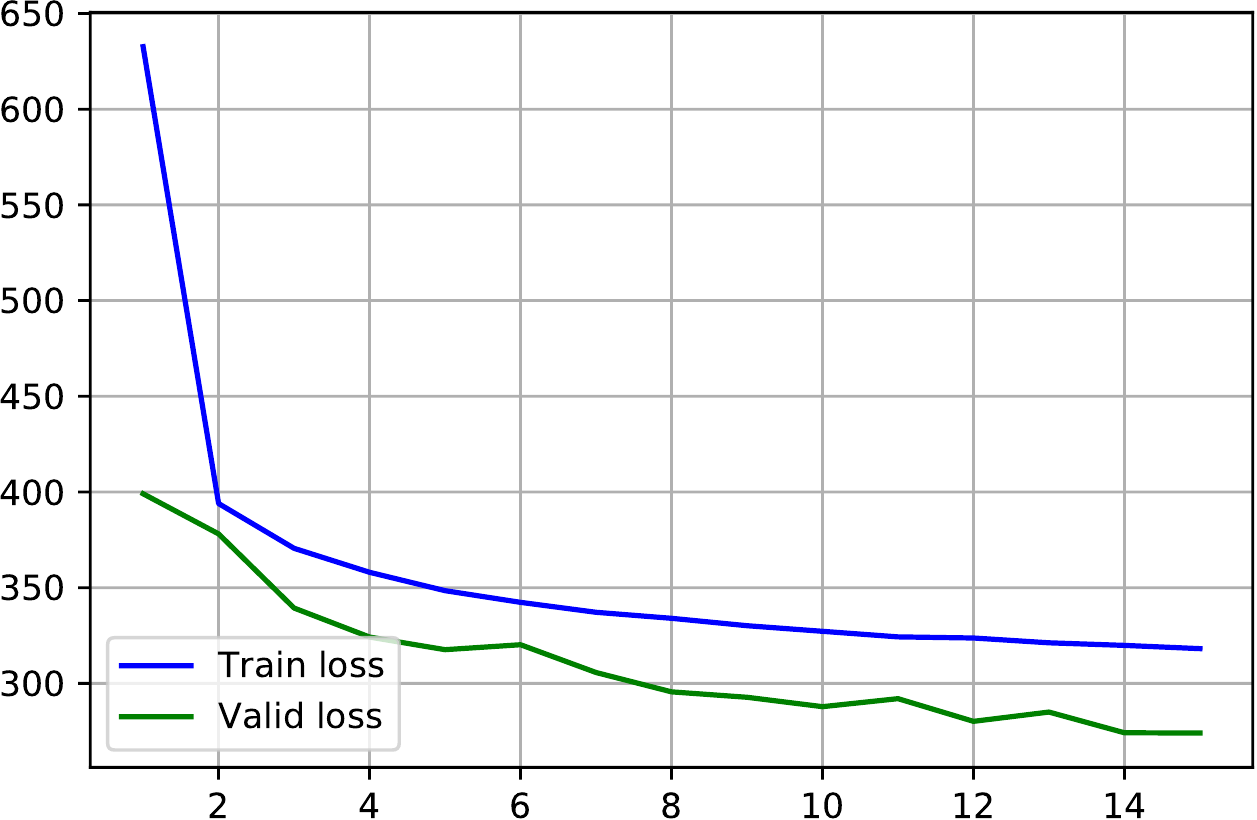}};
		\node[below=of img, node distance=0cm, yshift=1cm,font=\color{black}] {Epochs};
		\node[left=of img, node distance=0cm, rotate=90, anchor=center,yshift=-0.7cm,font=\color{black}] {MAE};
		\end{tikzpicture}
		\caption{Validation station 27.}
	\end{subfigure}%
	
	\vspace{15pt}
	
	\begin{subfigure}{0.48\textwidth}
		\centering
		\begin{tikzpicture}
		\node (img)  {\includegraphics[scale=0.55]{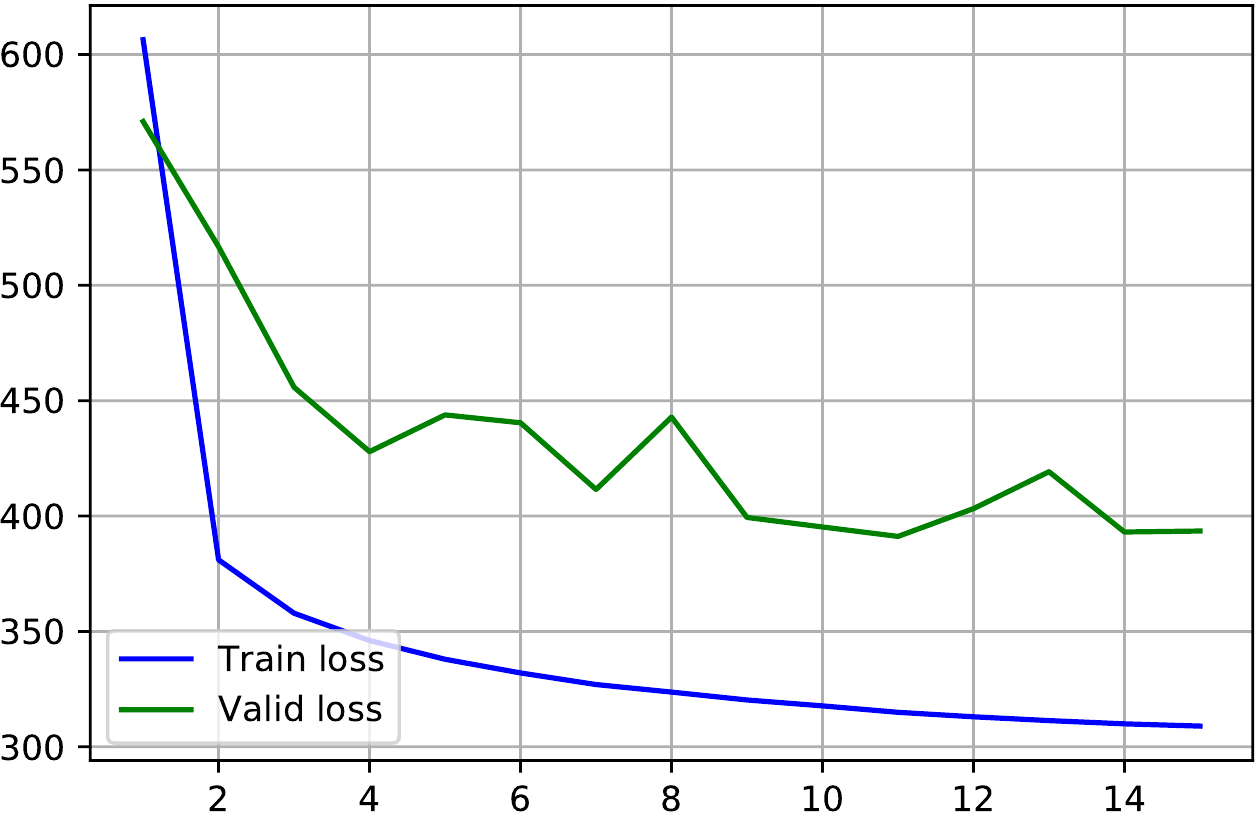}};
		\node[below=of img, node distance=0cm, yshift=1cm,font=\color{black}] {Epochs};
		\node[left=of img, node distance=0cm, rotate=90, anchor=center,yshift=-0.7cm,font=\color{black}] {MAE};
		\end{tikzpicture}
		\caption{Validation station 32.}
	\end{subfigure}\quad
	\begin{subfigure}{0.48\textwidth}
		\centering
		\begin{tikzpicture}
		\node (img)  {\includegraphics[scale=0.55]{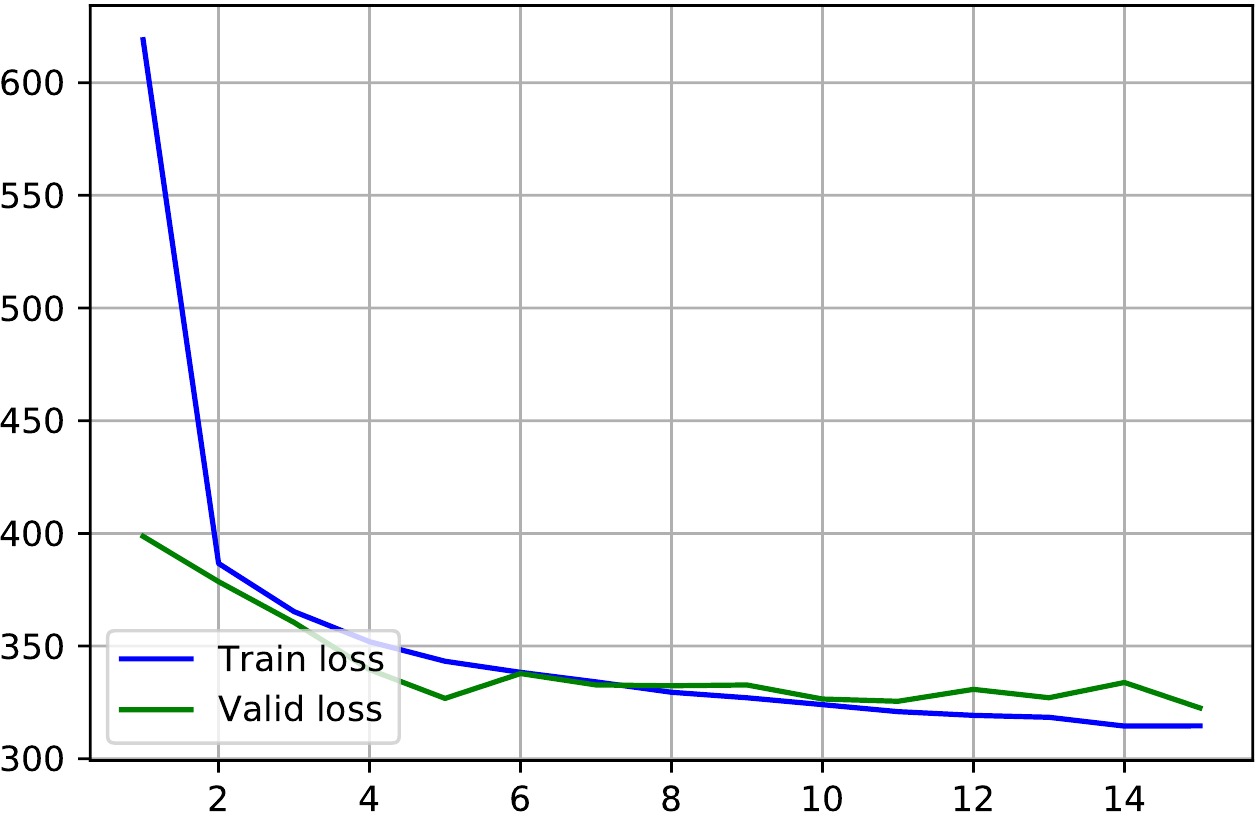}};
		\node[below=of img, node distance=0cm, yshift=1cm,font=\color{black}] {Epochs};
		\node[left=of img, node distance=0cm, rotate=90, anchor=center,yshift=-0.7cm,font=\color{black}] {MAE};
		\end{tikzpicture}
		\caption{Validation station 76.}\label{FigStation76}
	\end{subfigure}%
	\caption{Loss function values at 4 validation ATR locations do NOT indicate that the model overfits.}\label{LossPlots}
\end{figure}

Loss plots for the training and validation data at four ATR stations (Figure \ref{LossPlots}) show sound performance of the model. Specifically, we observe that loss function for both training and validation data shows a downward trend and then evens out with additional epochs. Most importantly, we do not observe the loss function for validation data exhibiting any significant upward trends with additional epochs, which would have suggested that the model overfits. Moreover, we provide some domain-specific explanations regarding observed values of the loss function. Namely, if traffic volumes at the ATR station used for validation corresponded very closely to the traffic volumes at the remaining stations used for training, then we would ideally expect to see that the loss function for training and validation data almost coincide. This seems to be the case when we use station 76 for validation (Figure \ref{FigStation76}). However, when we validate the model on station 14, which has much lower traffic volumes than all the other stations, then the loss function for validation data shows much smaller Mean Absolute Error (MAE) than its training counterpart (Figure \ref{FigStation14}). This is expected because MAE for a validation dataset with low average volume (e.g., 200 vehicles/hr) should be a lot smaller than in the case of a training dataset with high average volume (e.g., 1,000 vehicles/hr).

\subsection{Effect of dropout}
\begin{figure}
	\centering
	\begin{subfigure}{0.33\textwidth}
		\centering
		\includegraphics[height=33mm]{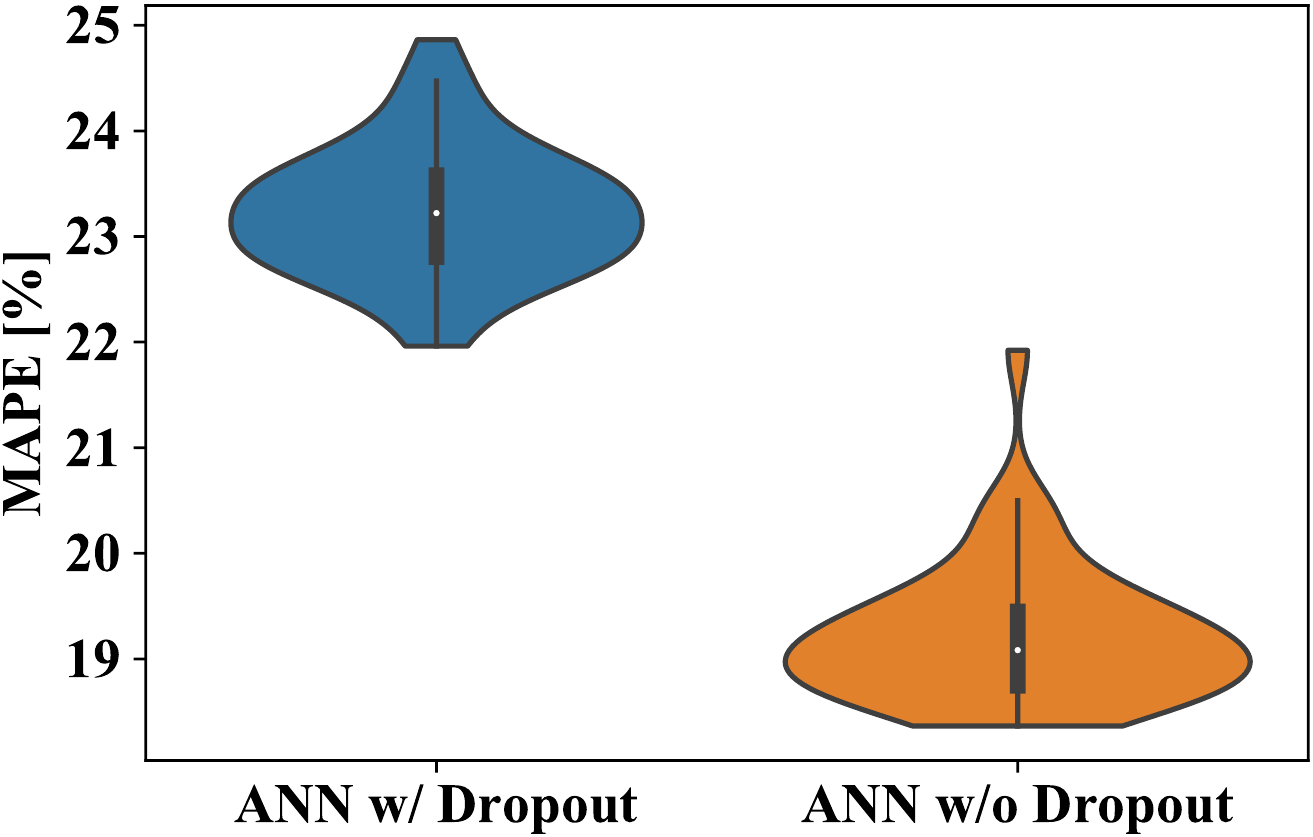}
		\caption{Training data.}
	\end{subfigure}%
	\begin{subfigure}{0.33\textwidth}
		\centering
		\includegraphics[height=33mm]{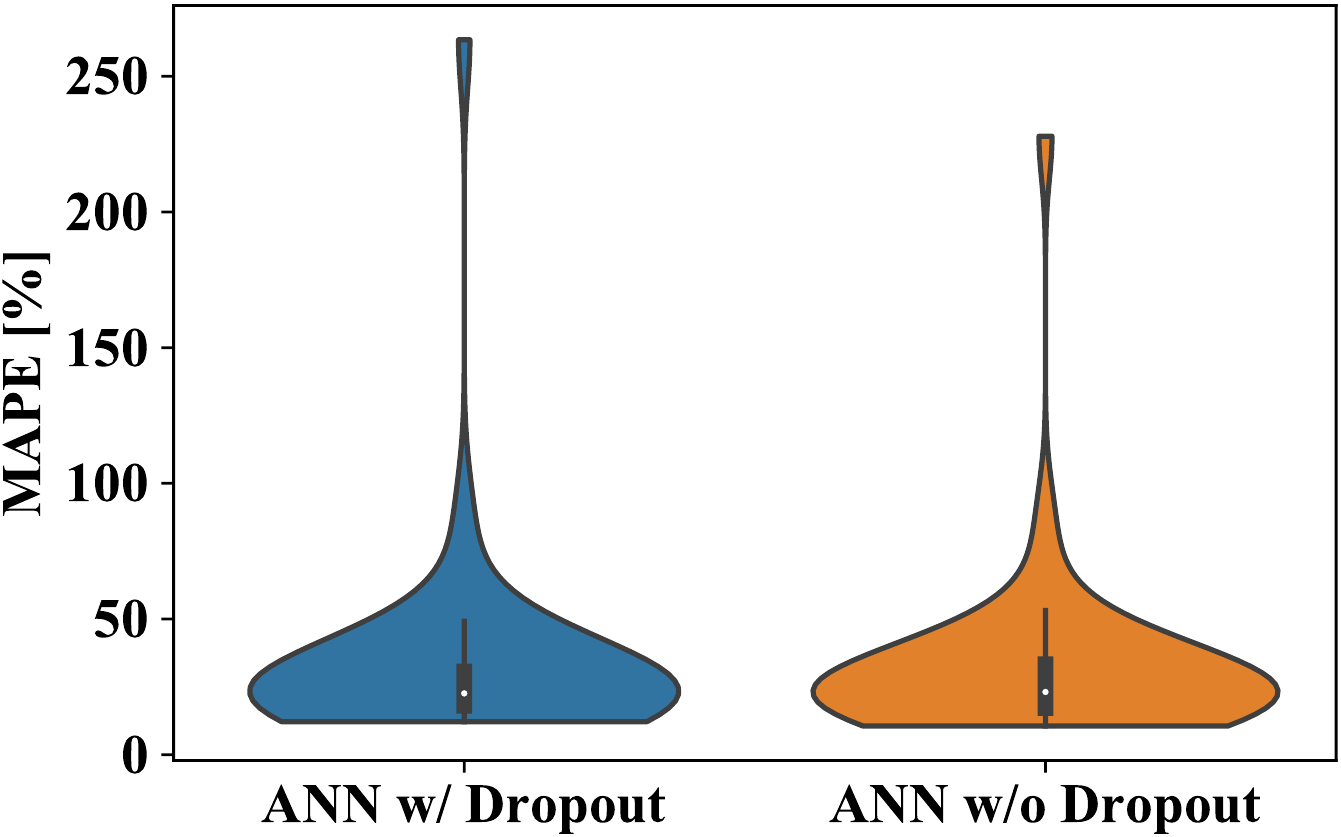}
		\caption{Test data.}
	\end{subfigure}%
	\begin{subfigure}{0.33\textwidth}
		\centering
		\includegraphics[height=33mm]{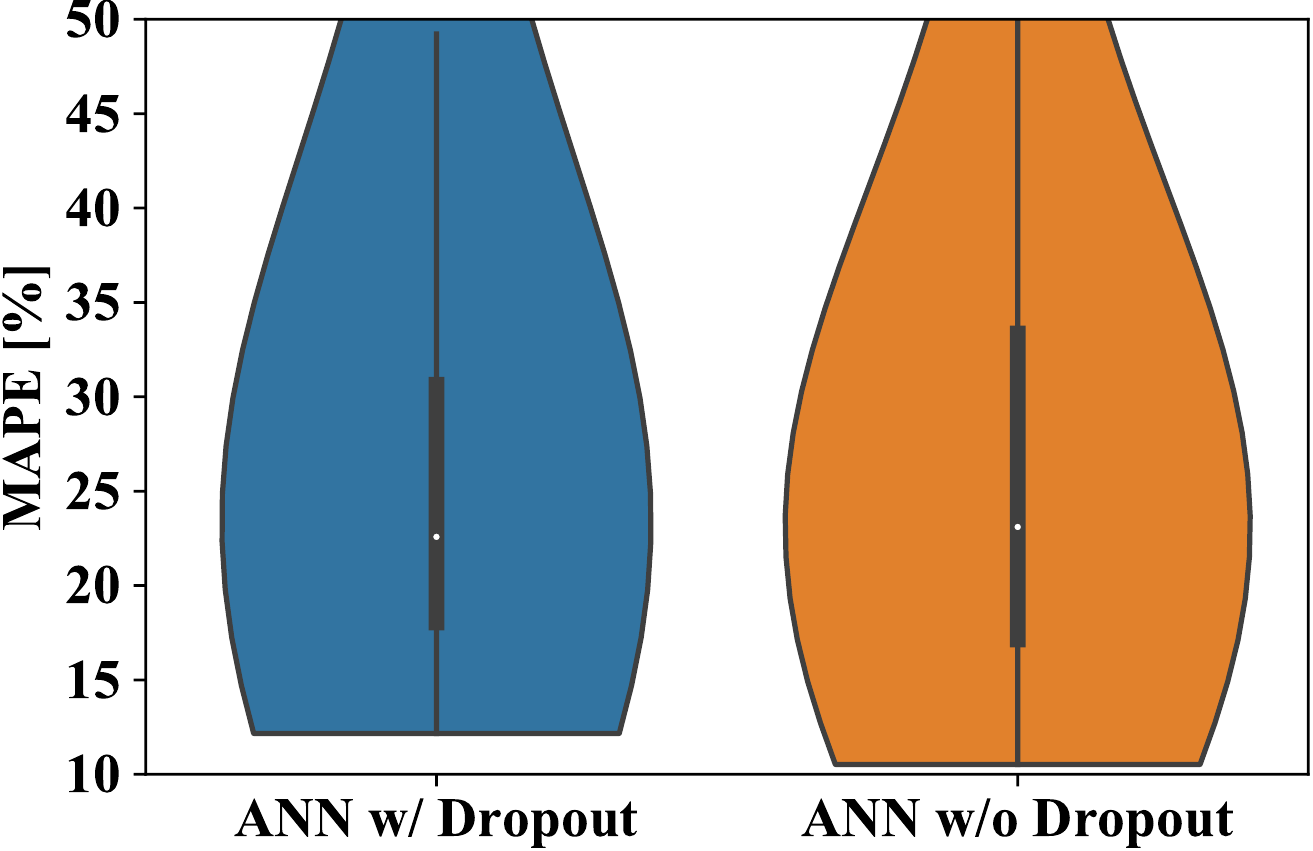}
		\caption{Test data, $\left[10,50\right]$ range.}
	\end{subfigure}
	\caption{Comparison of the model  with and without dropout shows that this procedure helps achieve almost identical median performance for the training and test data.}\label{DropoutPlots}
\end{figure}

We compared the performance of the model with and without applying the dropout procedure, and the corresponding results are summarized in Figure \ref{DropoutPlots}. It shows that not using dropout yields (notably) better results for training dataset (i.e., 23\% vs. 19\% median MAPE), and same results for the test data (i.e., 23\% median MAPE). In other words, Figure \ref{DropoutPlots} indicates that dropout helped achieve quite comparable results for the training and test data (i.e., median MAPE of about 23\% in both cases), which is what we would ideally like to observe. 

\end{document}